\documentclass[10pt,twocolumn,letterpaper]{article}

\usepackage{cvpr}
\usepackage{times}
\usepackage{epsfig}
\usepackage{graphicx}
\usepackage{subcaption}
\usepackage{amsmath}
\usepackage{amssymb}
\usepackage{float}
\usepackage{algorithm}
\usepackage[noend]{algpseudocode}
\usepackage{color,soul}

\usepackage[pagebackref=true,breaklinks=true,letterpaper=true,colorlinks,bookmarks=false]{hyperref}

\cvprfinalcopy 

\def\cvprPaperID{2331} 

\ifcvprfinal\fi

\title{RL-GAN-Net: A Reinforcement Learning Agent Controlled GAN Network for Real-Time Point Cloud Shape Completion }

\author{Muhammad Sarmad\\
KAIST\\
South Korea\\
{\tt\small sarmad@kaist.ac.kr}
\and
Hyunjoo Jenny Lee\thanks{co-corresponding authors}
\\
KAIST\\
South Korea\\
{\tt\small hyunjoo.lee@kaist.ac.kr}
\and
Young Min Kim\footnotemark[1]\\
KIST, SNU\\
South Korea\\
{\tt\small youngmin.kim@snu.ac.kr}
}

\begin{document}

\maketitle

\thispagestyle{empty}

\begin{abstract}
We present RL-GAN-Net, where a reinforcement learning (RL) agent provides fast and robust control of a generative adversarial network (GAN). Our framework is applied to point cloud shape completion that converts noisy, partial point cloud data into a high-fidelity completed shape by controlling the GAN.
While a GAN is unstable and hard to train, we circumvent the problem by (1) training the GAN on the latent space representation whose dimension is reduced compared to the raw point cloud input and (2) using an RL agent to find the correct input to the GAN to generate the latent space representation of the shape that best fits the current input of incomplete point cloud.
The suggested pipeline robustly completes point cloud with large missing regions. 
To the best of our knowledge, this is the first attempt to train an RL agent to control the GAN, which effectively learns the highly nonlinear mapping from the input noise of the GAN to the latent space of point cloud.
The RL agent replaces the need for complex optimization and consequently makes our technique real time. Additionally, we demonstrate that our pipelines can be used to enhance the classification accuracy of point cloud with missing data. 
\vspace{-2em}
\end{abstract}


\begin{figure}
  \centering
  \includegraphics[width=0.47\textwidth,trim={0 0.5cm 0 0},clip]{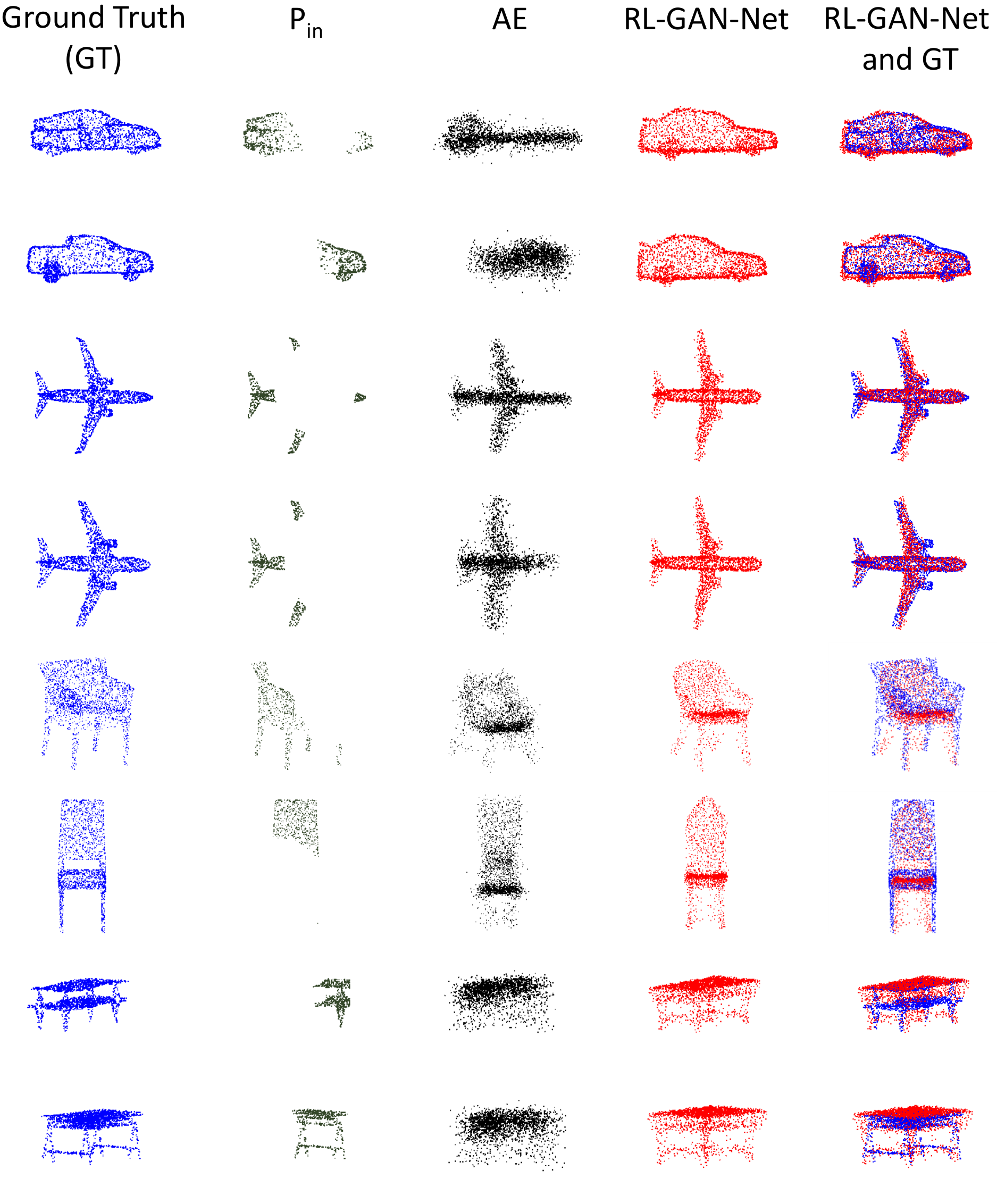}
  \caption{\small{{\bf Qualitative results of point cloud shape completion given input data missing 70\% of  its original points.} We present RL-GAN-Net, which observes a partial input point cloud data ($P_{in}$) and completes the shape within a matter of milliseconds. Even when input is severely distorted, our approach completes the shape with high-fidelity compared to the previous approach using autoencoder (AE)~\cite{panos}.}}\label{fig:results70}
\vspace{-2em}
\end{figure}

\begin{figure*}
    \centering
    \vspace{-0.5cm}
    \includegraphics[width=0.8\textwidth]{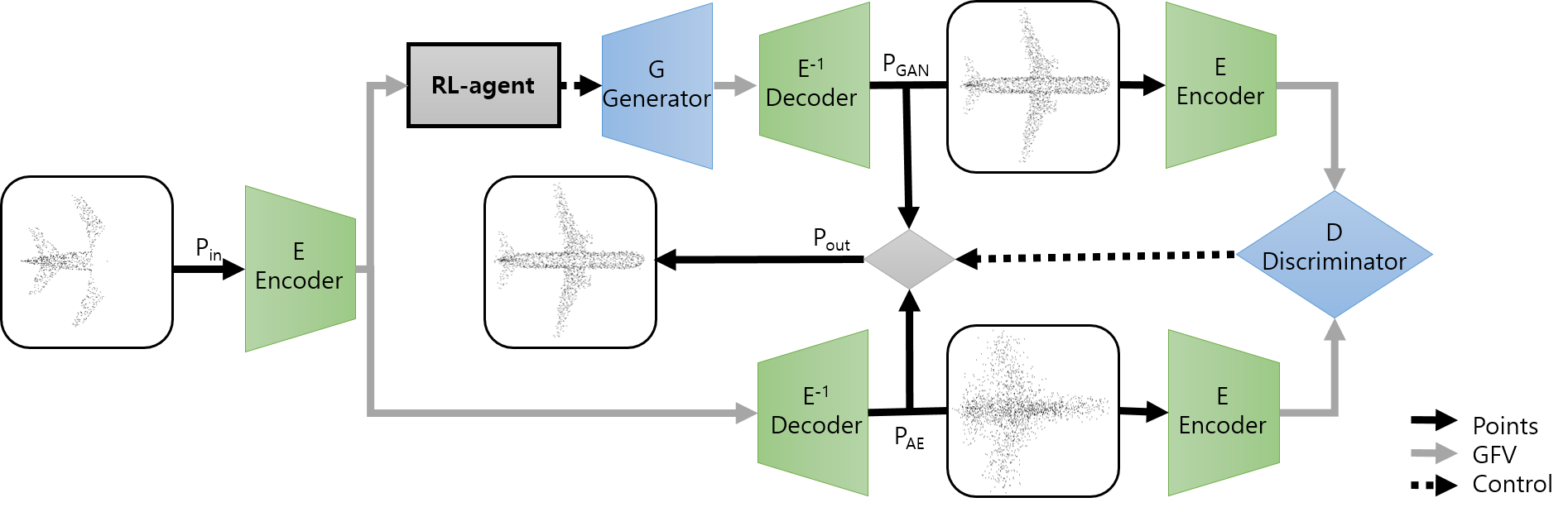}
    \caption{\small{{\bf The forward pass of our shape completion network.} By observing an encoded partial point cloud, our RL-GAN-Net selects an appropriate input for the latent GAN and generates a cleaned encoding for the shape. The synthesized latent representation is decoded to get the completed point cloud in real time. In our hybrid version, the discriminator finally selects the best shape. }}
    \label{fig:cover}
    \vspace{-1em}
\end{figure*}

\section{Introduction}

Acquisition of 3D data, either from laser scanners, stereo reconstruction, or RGB-D cameras, is in the form of the point cloud, which is a list of Cartesian coordinates.
The raw output usually suffers from large missing region due to limited viewing angles, occlusions, sensor resolution, or unstable measurement in the texture-less region (stereo reconstruction) or specular materials.
To utilize the measurements, further post-processing is essential which includes registration, denoising, resampling, semantic understanding and eventually reconstructing the 3D mesh model.

In this work, we focus on filling the missing regions within the 3D data by a data-driven method. The primary form of acquired measurements is the 3D point cloud which is unstructured and unordered. Therefore, it is not possible to directly apply conventional convolutional neural networks (CNN) approaches which work nicely for structured data e.g. for 2D grids of pixels~\cite{cnn1,cnn2,cnn3}. The extensions of CNN in 3D have been shown to work well with 3D voxel grid~\cite{PC2voxel,daiEPN,daiScan}. However, the computing cost grows drastically with voxel resolution due to the cubic nature of 3D space. Recently PointNet \cite{pointnet} has made it possible to directly process point cloud data despite its unstructured and permutation invariant nature. This has opened new avenues for employing point cloud data, instead of voxels, to contemporary computer-vision applications, e.g. segmentation, classification and shape completion~\cite{panos,CMUshape,pointnetplus,sonet,pointset}.

 In this paper, we propose our pipeline RL-GAN-Net as shown in Fig.~\ref{fig:cover}. It is a reinforcement learning agent controlled GAN (generative adversarial network) based network which can predict complete point cloud from incomplete data. 
As a pre-processing step, we train an autoencoder (AE) to get the latent space representation of the point cloud and we further use this representation to train a GAN~\cite{panos}. 
Our agent is then trained to take an 'action' by selecting an appropriate $z$ vector for the generator of the pre-trained GAN to synthesize the latent space representation of the complete point cloud.
Unlike the previous approaches which use back-propagation to find the correct $z$ vector of the GAN~\cite{CMUshape,inpainting}, our approach based on an RL agent is real time and also robust to large missing regions.
However, for data with small missing regions, a simple AE can reliably recover the original shape. Therefore, we use the help of a pre-trained discriminator of GAN to decide the winner between  the decoded output of the GAN and the output of the AE.
The final choice of completed shape  preserves the global structure of the shape and is consistent with the partial observation. A few results with 70\% missing data are shown in Fig.~\ref{fig:results70}.

To the best of our knowledge, we are the first to introduce this unique combination of RL and GAN for solving the point cloud shape completion problem.
We believe that the concept of using an RL agent to control the GAN's output opens up new possibilities to overcome underlying instabilities of current deep architectures. This can also lead to employing similar concept for problems that share the same fundamentals of shape completion e.g. image in-painting \cite{inpainting}.

Our key contributions are the following:
\begin{itemize}
\item We present a shape completion framework that is robust to low-availability of point cloud data without any prior knowledge about visibility or noise characteristics. 
\item We suggest a real-time control of GAN to quickly generate desired output without optimization. Because of the real-time nature, we demonstrate that our pipeline can pre-process the input for other point cloud processing pipelines, such as classification.
\item We demonstrate the first attempt to use deep RL framework for the shape completion problem. In doing so, we demonstrate a unique RL problem formulation.

\end{itemize}

\section{Related Works}

\paragraph{Shape Completion and Deep Learning.}
3D shape completion is a fundamental problem which is faced when processing 3D measurements of the real world.
Regardless of the modality of the sensors (multi-view stereo, the structure of light sensors, RGB-D cameras, lidars, etc.), the output point cloud exhibits large holes due to complex occlusions, limited field of view and unreliable measurements (because of material properties or texture-less regions). 
Early works use symmetry~\cite{symmetry} or example shapes~\cite{example} to fill the missing regions.
More recently databases of shapes has been used to retrieve the shape that is the closest to the current measurement~\cite{guidedScan2013,retrieval}.

Recently, deep learning has revolutionized the field of computer vision due to the enhanced computational power, the availability of large datasets, and the introduction of efficient architectures, such as the CNN~\cite{cnn3}.
Deep learning has demonstrated superior performance on many traditional computer vision tasks such as classification~\cite{cnn1,cnn2,res1} and segmentation~\cite{fcnSemantic,fcnSemantic2}.
Our 3D shape completion adapts the successful techniques from the field of deep learning and uses data-driven methods to complete the missing parts.

3D deep learning architecture largely depends on the choice of the 3D data representation, namely volumetric voxel grid, mesh, or point cloud.
 The extension of CNN in 3D works best with 3D voxel grids, which can be generated from point measurements with additional processing. Dai et al.~\cite{daiEPN} introduced a voxel-based shape completion framework which consists of a data-driven network and an analytic 3D shape synthesis technique. 
However, voxel-based techniques are limited in resolution because the network complexity and required computations increase drastically with the resolution. 
Recently, Dai et al.~\cite{daiScan} extended this work to perform scene completion and semantic segmentation using coarse-to-fine strategy and using sub-volumes.
There are also manifold-based deep learning approaches~\cite{graphdeeplearning} to analyze various characteristics of complete shapes, but these lines of work depend on the topological structure of the mesh. Such techniques are not compatible with point cloud.

Point cloud is the raw output of many acquisition techniques. It is more efficient compared to the voxel-based representation, which is required to fully cover the entire volume including large empty spaces. However, most of the successful deep learning architectures can not be deployed on point cloud data. Stutz et al.~\cite{PC2voxel} introduced a network which consumes incomplete point cloud but they use a pre-trained decoder to get a voxelized representation of the complete shape. 
Direct point cloud processing has been made possible recently due to the emergence of new architecture such as PointNet~\cite{pointnet} and others~\cite{pointnetplus,sonet,rsnet}.  

Achlioptas et al.~\cite{panos} explored learning shape representation with auto-encoder. They also investigated the generation of 3D point clouds and their latent representation with GANs.
Even though their work performs a certain level of shape completion, their architecture is not designed for shape completion tasks and suffers from considerable degradation as the number of missing points at the input are increased. 
Gurumurthy et al.~\cite{CMUshape} have suggested shape completion architecture which utilizes latent GAN and auto-encoder. However, they use a time-consuming optimization step for each batch of input to select the best seed for the GAN. While we also use latent GAN, our approach is different because we use a trained agent to find the GAN's input seed. In doing so, we complete shapes in a matter of milliseconds.
\paragraph{GAN and RL.}
Recently, Goodfellow et al.~\cite{gan1} suggested generative adversarial networks (GANs) which use a neural network (a discriminator) to train another neural network (a generator).
The generator tries to fool the discriminator by synthesizing fake examples that resembles real data, whereas the discriminator tries to discriminate between the real and the fake data. 
The two networks compete with each other and eventually the generator learns the distribution of the real data.

While GAN suggests a way to overcome the limitation of data-driven methods, at the same time, it is very hard to train and is susceptible to a local optimum. 
Many improvements have been suggested which range from changes in the architecture of generator and discriminator to modifications in the loss function and adoption of good training practices~\cite{wgan,selfgan,egan,wgangp}.
There are also practices to control GAN by observing the condition as an additional input~\cite{Mirza2014ConditionalGAN} or using back-propagation to minimize the loss between the desired output and the generated output~\cite{inpainting,CMUshape}.

Our pipeline utilizes deep reinforcement learning (RL) to control the complex latent space of GAN. 
RL is a framework where a decision-making network, also called an agent, interacts with the environment by taking available actions and collects rewards. 
RL agents in {\it discrete} action spaces have been used to provide useful guides to computer vision problems such as to propose bounding box locations~\cite{DQN_obj,DQN_obj2} or seed points for segmentation~\cite{DQN_seed} with deep Q-network (DQN)~\cite{dqn}.
On the other hand, we train an actor-critic based network~\cite{ddpg}
learning the policy in {\it continuous} action space to control GAN for shape completion.
In our setup, the environment is the shape completion framework composed of various blocks such as AE and GAN, and the action is the input to the generator.
The unknown behavior of the complex network can be controlled with the deep RL agent and we can generate completed shapes from highly occluded point cloud data.

\section{Methods}
\label{sec:methods}

Our shape completion pipeline is composed of three fundamental building blocks, which are namely an autoencoder (AE), a latent-space generative adversarial network ($l$-GAN) and a reinforcement learning (RL) agent. 
 Each of the components is a deep neural network that has to be trained separately. 
We first train an AE and use the encoded data to train $l$-GAN.
The RL agent is trained in combination with a pre-trained AE and GAN.

The forward pass of our method can be seen in Fig.~\ref{fig:cover}. The encoder of the trained AE encodes the noisy and incomplete point cloud to a noisy global feature vector (GFV). Given this noisy GFV, our trained RL agent selects the correct seed for the $l$-GAN's generator.
The generator produces the clean GFV which is finally passed through the decoder of AE to get the completed point cloud representation of the clean GFV.
A discriminator observes the GFV of the generated shape and the one processed by AE and selects the more plausible shape.
In the following subsections, we explain the three fundamental building blocks of our approach, then describe the combined architecture.

\subsection{Autoencoder (AE)}\label{sec:AE}

An AE creates a low-dimensional encoding of input data by training a network that reproduces the input.
An AE is composed of an encoder and a decoder. 
The encoder converts a complex input into an encoded representation, and the decoder reverts the encoded version back to the original dimension.
We refer to the efficient intermediate representation as the GFV
, which is obtained upon training an AE.
The training of AE is performed with back-propagation reducing the distance between input and output point cloud, either with the Earth Movers distance (EMD)~\cite{emd} or the Chamfer distance~\cite{pointset,panos}.  We use the Chamfer distance over EMD due to its efficiency  which can be defined as follows:
\begin{equation}
d_{CH} (P_1,P_2) = \sum_{a\in P_1} \min_{b\in P_2} \left\lVert a-b \right\rVert^2_2  + \sum_{b\in P_2} \min_{a\in P_1} \left\lVert a-b \right\rVert^2_2,
\label{eq:chamfer}
\end{equation}
where in Eq.~(\ref{eq:chamfer}) the $P_1$ and $P_2$ are the input and output point cloud respectively.

We first train a network similar to the one reported by Achlioptas et al.~\cite{panos} on the ShapeNet point cloud dataset \cite{modelnet,shapenet}. Achlioptas et al.~\cite{panos} also demonstrated that a trained AE can be used for shape completion. The trained decoder maps GFV into a complete point cloud even when the input GFV has been produced from an incomplete point cloud. But the performance degrades drastically as the percentage of the missing data in the input is increased (Fig.~\ref{fig:results70}).

\subsection{$l$-GAN}\label{sec:GAN}

GAN generates new yet realistic data by jointly training a pair of generator and discriminator~\cite{gan1}. While GAN demonstrated its success in image generation tasks \cite{selfgan,wgangp,wgan}, in practice, training a GAN tends to be unstable and suffer from mode collapse \cite{equalgan}.
Achlioptas et al.~\cite{panos} showed that training a GAN on GFV, or latent representation, leads to more stable training results compared to training on raw point clouds.
Similarly, we also train a GAN on GFV, which has been converted from complete point cloud data using the encoder of trained AE, Sec.~\ref{sec:AE}.
The generator synthesizes a new GFV from a noise seed $z$, which can then be converted into a complete 3D point cloud using the decoder of AE.
We refer to the network as $l$-GAN or latent-GAN. 

Gurumurthy et al.~\cite{CMUshape} similarly utilized $l$-GAN for point cloud shape completion.
They formulated an optimization framework to find the best input $z$ to the generator to create GFV that best explains the incomplete point cloud at the input.
However, as the mapping between the raw points and the GFV is highly non-linear, the optimization could not be written as a simple back-propagation.
Rather, the energy term is a combination of three loss terms.
We list the losses below, where $P_{in}$ is the incomplete point cloud input, $E$ and $E^{-1}$ are the encoder and the decoder of AE, and $G$ and $D$ represent the generator and the discriminator of the $l$-GAN respectively.

\begin{itemize}
\item Chamfer loss: the Chamfer distance between the input partial pointcloud $P_{in}$ and the generated, decoded pointcloud $E^{-1}(G(z))$
\begin{equation}
L_{CH}  = d_{CH}(P_{in}, E^{-1}(G(z)))
\label{eq:L_CH}
\end{equation}
\vspace{-2em}

\item GFV loss: $l_2$ distance between the generated GFV $G(z)$ and the GFV of the input pointcloud $E(P_{in})$
\begin{equation}
L_{GFV} =  \left\lVert G(z)-E(P_{in}) \right\rVert^2_2
\label{eq:L_GFV}
\end{equation}
\vspace{-2em}

\item Discriminator loss: the output of the discriminator
\begin{equation}
L_{D}  = -D(G(z))
\label{eq:rD}
\end{equation}
\vspace{-2em}

\end{itemize}

Gurumurthy et al.~\cite{CMUshape} optimized the energy function defined as a weighted sum of the losses, and the weights gradually evolve with every iteration.
However, we propose a more robust control of GAN using an RL framework, where an RL agent quickly finds the $z$-input to the GAN by observing the combination of losses.

\subsection{Reinforcement Learning (RL)}

In a typical RL-based framework, an agent acts in an environment
. Given an observation $x_t$ at each time step $t$, the agent performs an action $a_t$ and receives a reward $r_t$. The agent network learns a policy $\pi$ which maps states to the action with some probability. The environment can be modeled as a Markov decision process, i.e., the current state and action only depend on the previous state and action. 
The reward at any given state is the discounted future reward $R_t= \sum^T_{i=t}\gamma^{(i-t)}r(s_i,a_i)$.
The final objective is to find a policy which provides the maximum reward.

\begin{figure}
\begin{center}
   \includegraphics[width=0.47\textwidth]{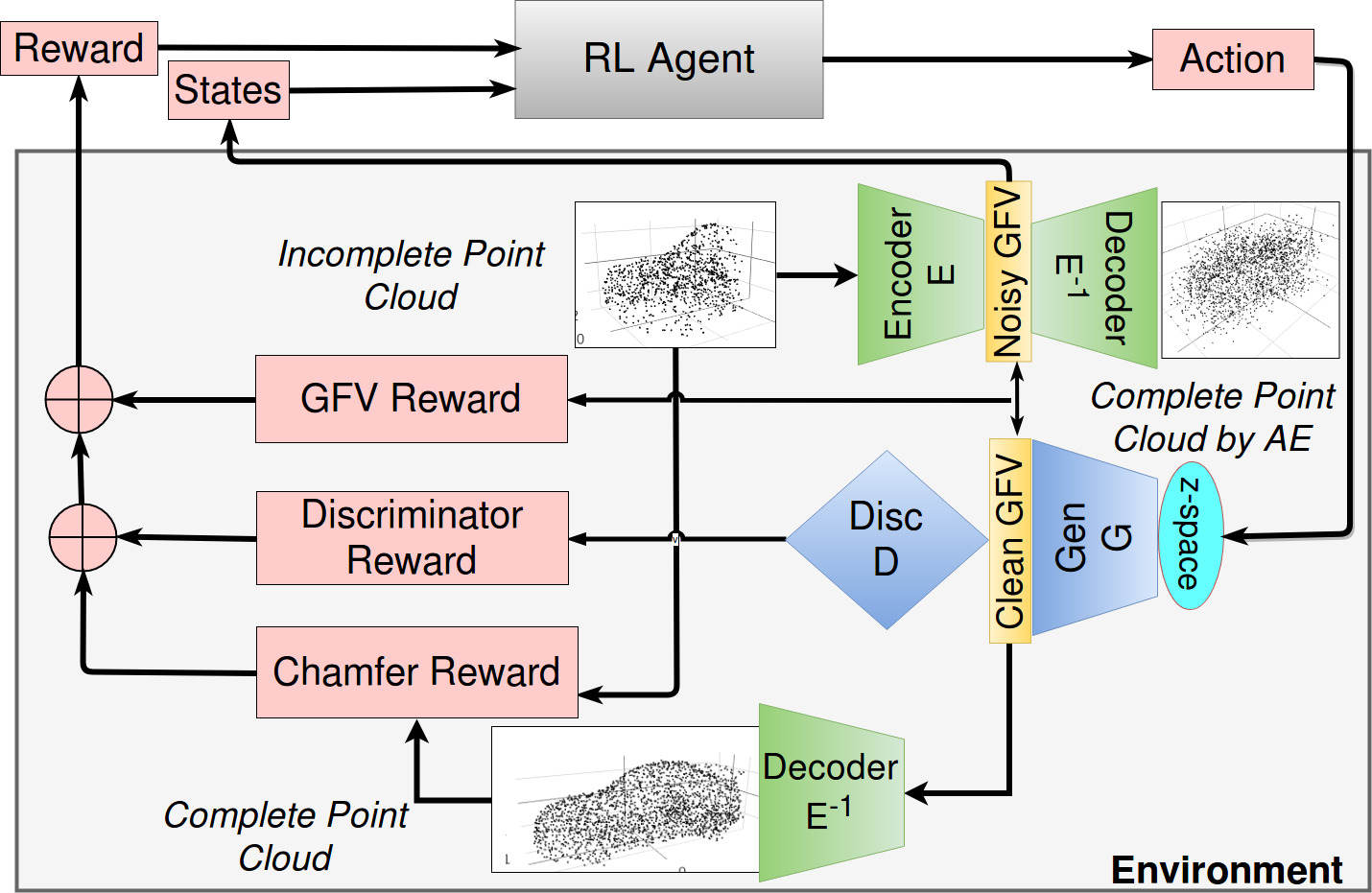}
\end{center}
   \caption{\small{{\bf Training RL-GAN-Net for shape completion.} Our RL framework utilizes AE (shown in green) and $l$-GAN (shown in blue). The RL agent and the environment are shaded in gray, and the embedded reward, states, and action spaces are highlighted in red. The output is decoded and completed as shown at the bottom. Note that the decoder and decoded point cloud in the upper right corner is added for a comparison, and does not affect the training. By employing an RL agent, our pipeline is capable of real-time shape completion.}}
\label{fig:rlnet}
\vspace{-1em}
\end{figure}

We formulate the shape completion task in an RL framework as shown in Fig.~\ref{fig:rlnet}.
For our problem, the environment is the combination of AE and $l$-GAN, and resulting losses that are calculated as intermediate results of various networks in addition to the discrepancy between the input and the predicted shape.
The observed state $s_t$ is the initial noisy GFV encoded from the incomplete input point cloud.
We assume that the environment is Markov and fully observed; i.e., the recent most observation $x_t$ is enough to define the state $s_t$.
The agent takes an action $a_t$ to pick the correct seed for the $z$-space input of the generator.
The synthesized GFV is then passed through the decoder to obtain the completed point cloud shape. 

One of the major tasks in training an RL agent is the correct formulation of the reward function.
Depending on the quality of the action, the environment gives a reward $r$ back to the agent.
In RL-GAN-Net, the right decision equates to the correct seed selection for the generator.
We use the combination of negated loss functions as a reward for shape completion task~\cite{CMUshape} (Sec.~\ref{sec:GAN}) that represent losses in all of Cartesian coordinate ($r_{CH}=-L_{CH}$), latent space ($r_{GFV}=-L_{GFV}$), and in the view of the discriminator ($r_D=-L_D$). The final reward term is given as follows:
\begin{equation}
r = w_{CH} \cdot r_{CH} + w_{GFV} \cdot r_{GFV} + w_D \cdot r_D,
\label{eq:reward}
\end{equation}
where $w_{CH}$, $w_{GFV}$, and $w_D$ are the corresponding weights assigned to each loss function. We explain the selection of weights in the supplementary material.

Since the action space is continuous, we adopt deep deterministic policy gradient (DDPG) by Lillicrap et al.~\cite{ddpg}. 
In DDPG algorithm, a parameterized actor network $\mu(s\mid\theta^\mu)$ learns a particular policy and maps states to particular actions in a deterministic manner. The critic network $Q(s,a)$ uses the Bellman equation and provides a measure of the quality of action and the state.
The actor network is trained by finding the expected return of the gradient to the cost $J$ $w.r.t$ the actor-network parameters, which is also known as the $policy$ $gradient$. It can be defined as below:
\begin{equation}
\begin{aligned}
\bigtriangledown_{\theta^\mu} J(\theta)=\mathbb{E}_{s_t\sim\rho^\beta}[\bigtriangledown_\alpha Q(s,a\mid\theta^Q)\mid_{s=s_t,a=\mu(s_t)} \\\bigtriangledown_{\theta_\mu}\mu(s\mid\theta^\mu)\mid_{s=s_t}]
\end{aligned}
\end{equation}

Before training the agent, we make sure that AE and GAN are adequately pre-trained as they constitute the environment. The agent relies on them to select the correct action.  
The algorithm of the detailed training process is summarized in Algorithm~\ref{algo:algo1}. 

\begin{algorithm}
\caption{Training RL-GAN-Net}\label{euclid}

\textbf{Agent Input:}

\begin{algorithmic}
\State State $(s_t)$: $s_t=GFV_n$ = \textbf{E}($P_{in}$); Sample pointcloud \textbf{$P_{in}$} from dataset into the pre-trained encoder \textbf{E} to generate noisy latent representation \textbf{$GFV_n$}.
\State Reward $(r_t)$: Calculated using Eq. (\ref{eq:reward})
\end{algorithmic}

\textbf{Agent Output:}

\begin{algorithmic}
\State Action $(a_t)$: $a_t=z$
\State Pass $z$-vector to the pre-trained generator \textbf{G} to form clean latent vector \textbf{$GFV_c$}=\textbf{G}($z$)
\end{algorithmic}

\textbf{Final Output:}
\begin{algorithmic}
\State $P_{out}$ = \textbf{E$^{-1}(GFV_c)$}; Pass $GFV_c$ into decoder E$^{-1}$ to generate output point cloud $P_{out}$.
\end{algorithmic}

\begin{algorithmic}[1]
\State Initialize \textbf{procedure Env} with pre-trained generator \textbf{G}, discriminator \textbf{D}, encoder \textbf{E} and decoder \textbf{E$^{-1}$}

\State Initialize policy $\pi$ with \textbf{DDPG}, actor \textbf{A}, critic \textbf{C}, and 

replay buffer \textbf{R}

\For{$t_{steps}$ $<$ $max steps$}

\State Get $P_{in}$

\If {$t_{steps} > 0$}
\State Train \textbf{A} and \textbf{C} with \textbf{R}
\EndIf

\If {$t_{Last Evaluation} >f_{Eval Frequency}$}
\State Evaluate $\pi$
\EndIf

\State $GFV_n \gets$ \textbf{E}($P_{in}$)
\If {$t_{steps} > t_{StartTime}$}
\State Random Action $a_t$
\EndIf
\If {$t_{steps} < t_{StartTime}$}
\State  Use $a_t \gets$ \textbf{A} $\gets GFV_n$
\EndIf

\State  $(s_t,a_t,r_t,s_{t+1})\gets$\textbf{Env} $\gets$$a_t$

\State Store transition $(s_t,a_t,r_t,s_{t+1})$ in \textbf{R}
\EndFor{\textbf{endfor}}
\Procedure{\textbf{Env}}{$P_{in}$,$a_t$} 

\State Get State ($s_t$) : $GFV_n$ $\gets$ \textbf{E}($P_{in}$)


\State Implement Action : $GFV_c \gets$ \textbf{G} $(a_t=z)$

\State Calculate reward $r_t$ using Eq.~(\ref{eq:reward})

\State Obtain point cloud : $P_{out} \gets$ \textbf{E$^{-1}$} $(GFV_c)$

\EndProcedure
\end{algorithmic}
\label{algo:algo1}
\end{algorithm}

\subsection{Hybrid RL-GAN-Net}

With the vanilla implementation described above, the generated details of completed point cloud can sometimes have limited semantic variations.
When the portion of missing data is relatively small, the AE can often complete the shape that agrees better with the input point cloud.
On the other hand, the performance of AE degrades significantly as more data is missing, and our RL-agent can nonetheless find the correct semantic shape. 
Based on this observation, we suggest a hybrid approach  by using a discriminator as  a switch that selects the best results out of the vanilla RL-GAN-Net and AE.
The final pipeline we used for the result is shown in Fig.~\ref{fig:cover}.
Our hybrid approach can robustly complete the semantic shape in real time and at the same time preserve local details.

\section{Experiments}

We used PyTorch \cite{pytorch} and open source codes \cite{ddpgcode,gitSO,gitgan,gitddpg} for our implementation. All networks were trained on a single Nvidia GTX Titan Xp graphics card. The details of network architectures are provided in the supplementary materials.
%
For the experiments, we used the four categories with the most number of shapes among ShapeNetCore~\cite{shapenet,panos} dataset, namely cars, airplanes, chairs, and desks.
The total number of shapes sums to 26,829 for the four classes.
All shapes are translated to be centered at the origin and scaled such that the diagonals of bounding boxes are of unit length. 
The ground-truth point cloud data is generated by uniformly sampling 2048 points on each shape.
The points are used to train the AE and generate clean GFV to train the $l$-GAN.
The incomplete point cloud is generated by selecting a random seed from the complete point cloud and removing points within a certain radius. 
The radius is controlled for each shape to obtain the desired amount of missing data.
We generated incomplete point cloud missing 20, 30, 40, 50 and 70\% of the original data for test, and trained our RL agent on the complete dataset.

\begin{figure}
  \centering
  \includegraphics[width=0.47\textwidth,trim={0 0.5cm 0 0},clip]{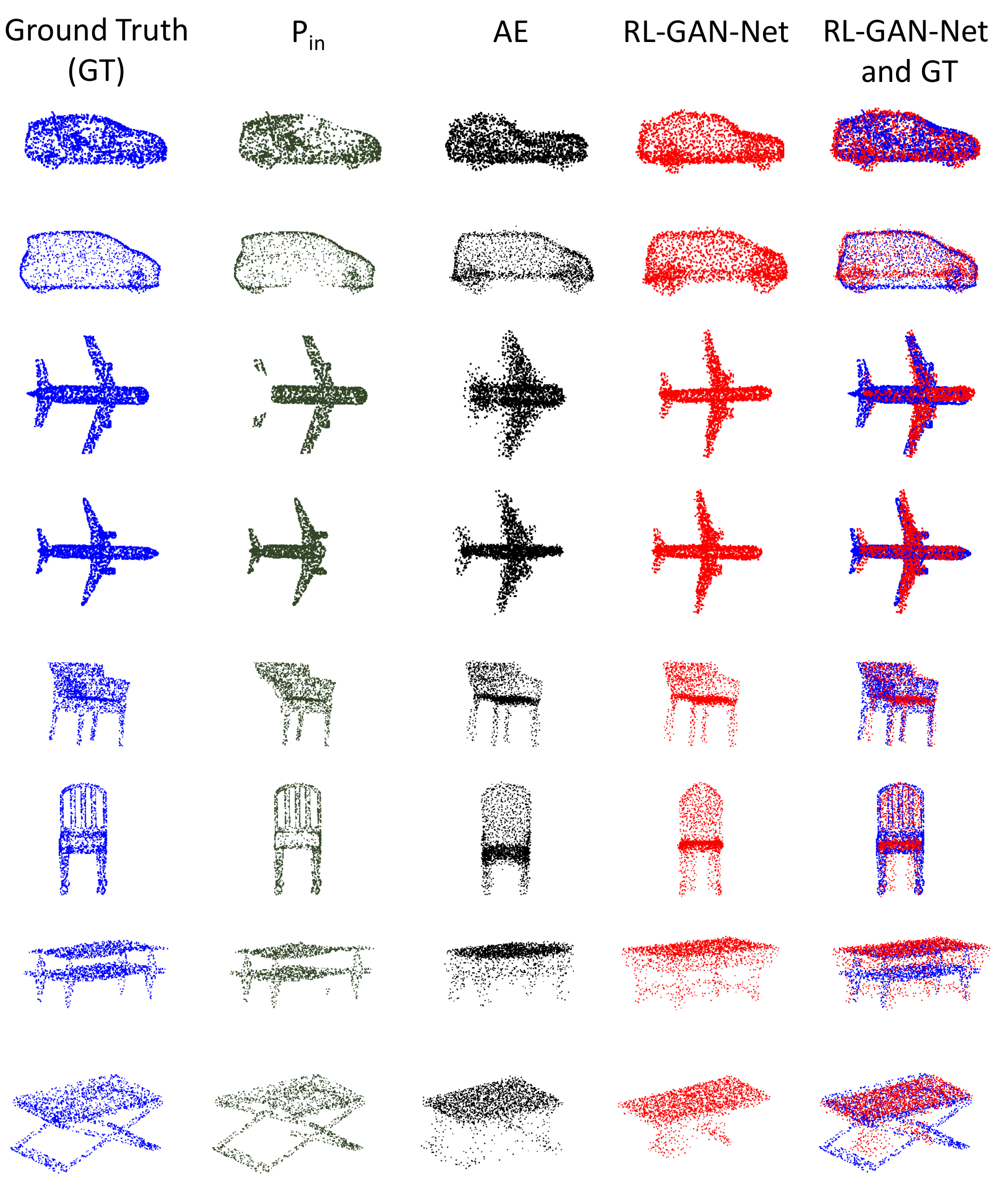}
  \caption{\small{{\bf Qualitative results of point cloud shape completion missing 20\% of its original points.} With relatively small missing data, AE sometimes performs better in completing shapes. Therefore, our hybrid RL-GAN-Net reliably selects the best output shape among the AE and the vanilla RL-GAN-Net.}}\label{fig:results20}
  \vspace{-1em}
\end{figure}

\begin{figure*}
    \vspace{-0.35cm}
    \centering
    \begin{subfigure}[b]{0.3\textwidth}
        \includegraphics[width=\textwidth]{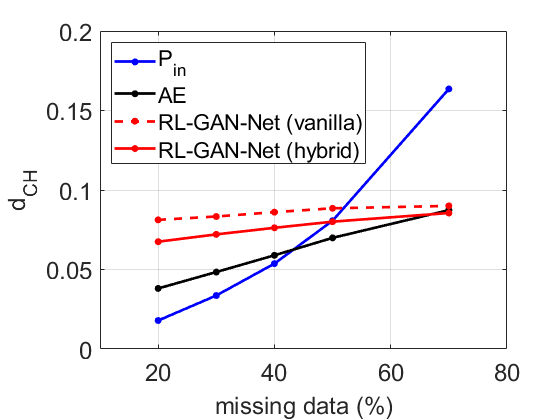}
        \caption{\small{Chamfer distance to GT}}
        \label{fig:AE_RL}
    \end{subfigure}
    ~
     \begin{subfigure}[b]{0.3\textwidth}
        \includegraphics[width=\textwidth]{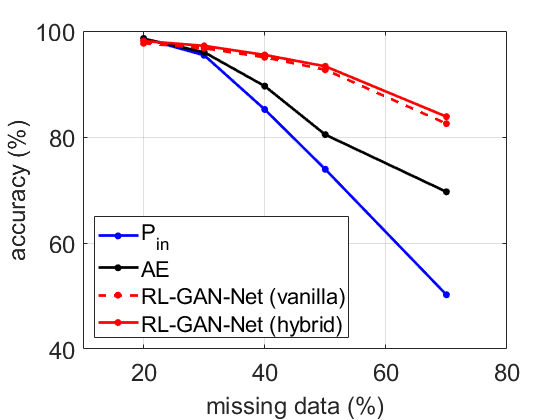}
        \caption{\small{Classification accuracy~\cite{pointnet}}}
        \label{fig:classification}
    \end{subfigure}
    ~ 
    \begin{subfigure}[b]{0.3\textwidth}
        \includegraphics[width=\textwidth]{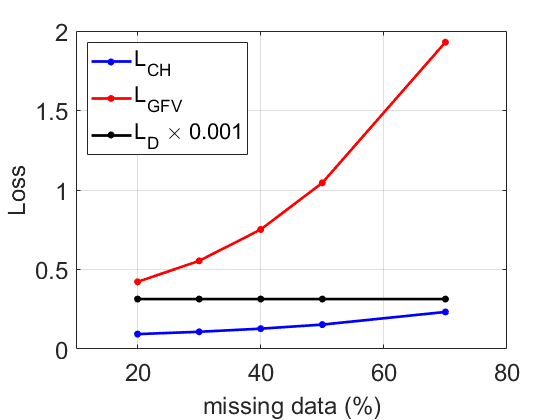}
        \caption{\small{Loss terms}}
        \label{fig:loss_graph}
    \end{subfigure}
    \caption{\small{{\bf Performance analysis.} We compare the two versions of our algorithms against the original input and the AE in terms of (a) the Chamfer distance (the lower the better) and (b) the performance gain for shape classification (the higher the better). (c) We also analyze the losses of RL-GAN-Net with different amount of missing data.}}\label{fig:graph}
    \vspace{-0.3cm}
\end{figure*}

\subsection{Shape Completion Results}
We present the results using the two variations of our algorithm, the vanilla version  and the hybrid approach as mentioned in Sec.~\ref{sec:methods}.
Since the area is relatively new, there are not many previous works available performing shape completion in point cloud space.
We compare our result against the method using AE only~\cite{panos}.

Fig.~\ref{fig:AE_RL} shows the Chamfer distances of the completed shape compared against the ground-truth point cloud.
With point cloud input with 70\% of its original points missing, the Chamfer distance compared to ground truth increase up to 16\% of the diagonal of the shape, but the reconstructed shapes of AE, vanilla and hybrid RL-GAN-Net all show less than 9\% of the distances. 

While the Chamfer distance is a widely used metric to compare shapes, we noticed that it might not be the absolute measure of the performance.
From Fig.~\ref{fig:AE_RL}, we noticed that the input point cloud $P_{in}$ was the best in terms of Chamfer distance for 20\% missing data.
However, from our visual inspection in Fig.~\ref{fig:results20}, the completed shapes, while they might not be exactly aligned in every detail, are semantically reasonable and does not exhibit any large holes that are clearly visible in the input.
For the examples with data missing 70\% of its original points in Fig.~\ref{fig:results70}, it is  obvious that our approach is superior to AE, whose visual quality for completed shape severely degrades as the ratio of missing data increases.
However, the Chamfer distance is almost the same for AE and RL-GAN-Net.
The observation can be interpreted as the fact that 1) AE is specifically trained to reduce the Chamfer loss, thus performs better in terms of the particular loss, while RL-GAN-Net jointly considers Chamfer loss, latent space and discriminator losses, and 2) $P_{in}$ has points that are exactly aligned with the GT, which, when averaged, compensates errors from missing regions.

Nonetheless our hybrid approach correctly predicts the category of shapes and  fills the missing points even with a large amount of missing data.
In addition, the RL-controlled forward pass takes only around a millisecond to complete, which is a huge advantage over previous work~\cite{CMUshape} that requires back-propagation over a complex network.
They claim the running time of 324 seconds for a batch of 50 shapes.
On the other hand, our approach is real-time and easily used as a preprocessing step for various tasks, even at the scanning stage.

\begin{table}
\centering
\begin{tabular}{|c|ccccc|}
  \hline
  ratio (\%) &  20 &  40 & 30 &  50 &  70\\
  \hline
  time  (ms) & 1.310 & 1.293 & 1.295 & 1.266 & 1.032 \\
  \hline
\end{tabular}
  \caption{\small{The average action time for the RL agent to produce clean GFV from observation of noisy GFV. Our approach can create the appropriate $z$-vector approximately in {\bf one millisecond}. }}\label{tb:timing}
  \vspace{-1em}
\end{table}

 \vspace{-1.0em}
\paragraph{Comparison with Dai et al.~\cite{daiEPN}}
While there is not much prior work in point cloud space, we include the completion results of Dai et al~\cite{daiEPN} which works in a different domain (voxel grid).
To briefly describe, their approach used an encoder-decoder network in $32^3$ voxel space followed by an analytic patch-based completion in $128^3$ resolution.
Their results of both resolutions are available as distance function format.
We converted the distance function into a surface representation using the MATLAB function {\it isosurface} as they described, and uniformly sampled 2048 points to compare with our results. We present the qualitative visual comparison in Fig.~\ref{fig:angela}.
The results of encoder-decoder based network (referred as Voxel $32^3$ in the figure) are smoother than point clouds processed by AE as the volume accumulation compensates for random noise. However, the approach is limited in resolution and washes out the local details. 
Even after the patch-based synthesis in $128^3$ resolution, the details they could recover are limited.
On the other hand, our approach robustly preserves semantic symmetries and completes local details in challenging scenarios.
It should be noted that we used only scanned point data but did not incorporate the additional mask information, which they utilized.
More results are included in the supplementary material due to the limitation of space.

\begin{figure}
  \centering
  \includegraphics[width=0.47\textwidth]{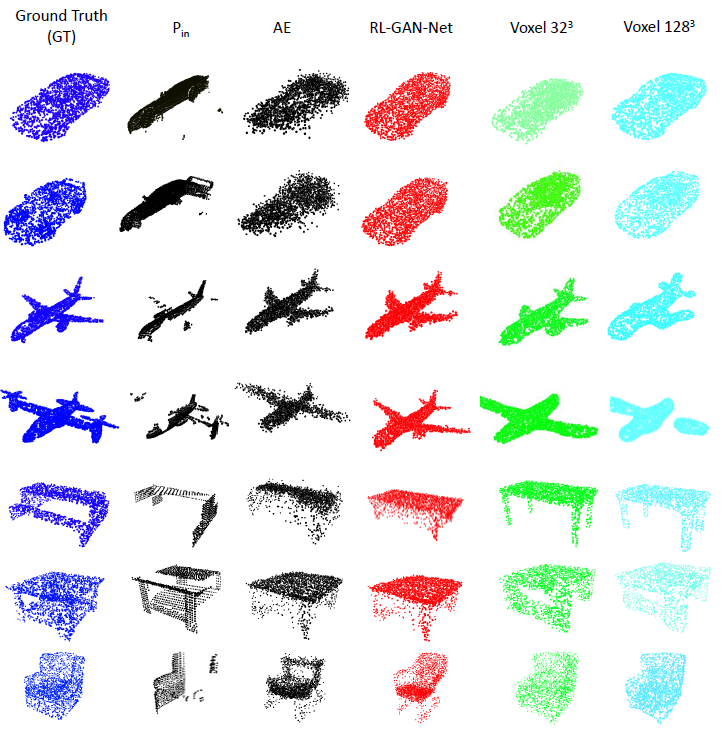}
  \caption{\small{\textbf{Performance Comparison.} Comparison of RLGAN-Net vs Dai et al.\cite{daiEPN} for their $32^3$ and $128^3$ resolution results. We converted their distance function output to point cloud domain. It should be noted that they additionally have mask  information whereas we operate directly on the scanned points only.}}\label{fig:angela}
  \vspace{-1em}
\end{figure}

\subsection{Application into Classification}
\label{sec:classification}

As an alternative measure to test the performances of the semantic shape completion, we compared the classification accuracy of $P_{in}$ and the shapes completed by AE and RL-GAN-Net.
This scenario also agrees with the main applications that we intended.
That is, RL-GAN-Net can be used as a quick preprocessing of the captured real data before performing other tasks as the raw output of 3D measurements are often partial, noisy data to be used as a direct input to point cloud processing framework.
We took the incomplete input and first processed through our shape completion pipeline.
Then we analyzed the classification accuracy of PointNet~\cite{pointnet} with the completed point cloud input and compared against the results with incomplete input.
Fig.~\ref{fig:classification} shows the improvement of classification accuracy.
Clearly, our suggested pipeline reduces possible performance losses of existing networks by completing the defects in the input.

\begin{figure}
  \centering
  \includegraphics[width=0.47\textwidth,trim={0 0.8cm 0 0}]{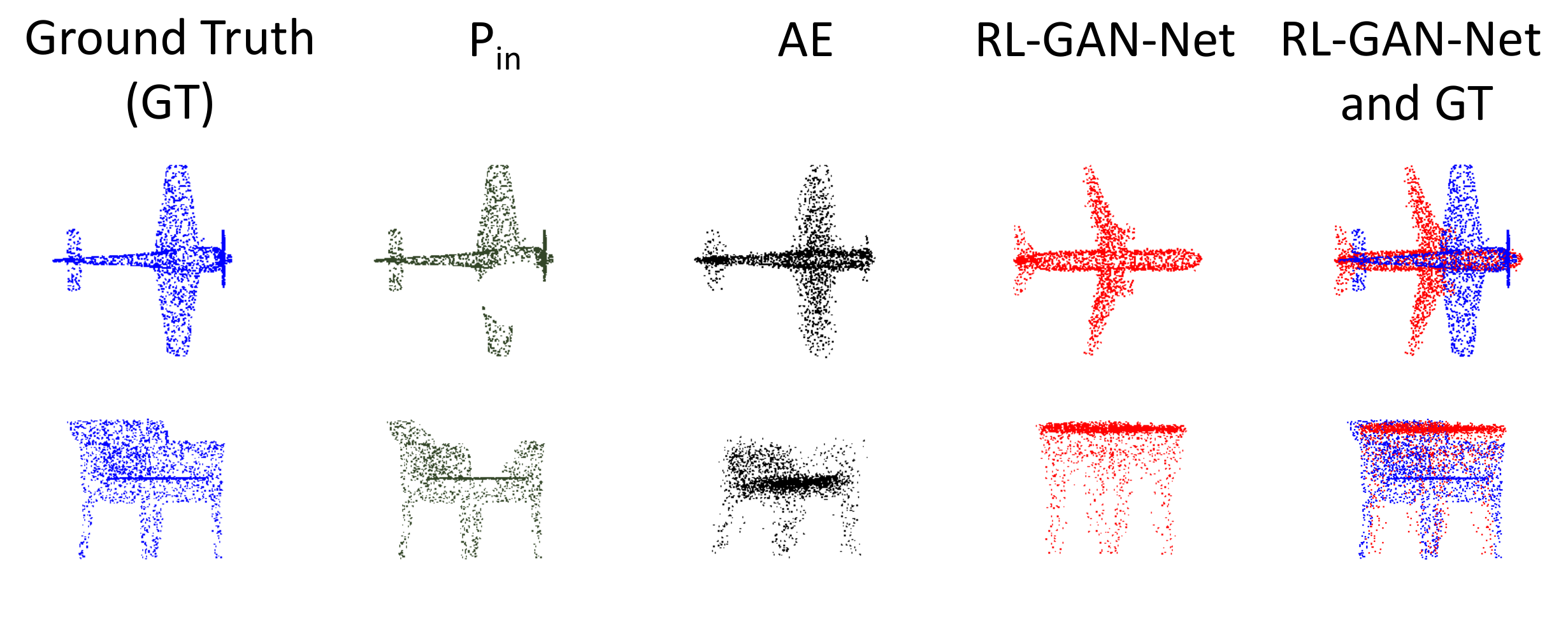}
  \caption{\small{{\bf Failure cases.} RL-GAN-Net can sometimes predict a wrong category (top) or semantically similar but different shape of the category (bottom).}}\label{fig:failure}
  \vspace{-1em}
\end{figure}

We also would like to add a note about the performance of the vanilla RL-GAN-Net and the hybrid approach. 
We noticed that the main achievement of our RL agent is often limited to finding the correct semantic categories in the latent space.
The hybrid approach overcomes the limitation by selecting the results of AE when the shape is more reasonable according to the trained discriminator.
This agrees with the fact that the hybrid approach is clearly better in terms of Chamfer distance in Fig.~\ref{fig:AE_RL}, but is comparable with the vanilla approach in classification in Fig.~\ref{fig:classification}, where the task is finding the correct category.
Fig.~\ref{fig:failure} shows some examples of failure cases, where the suggested internal category does not exactly align with the observed shape.

\subsection{Reward Function Analysis}
We demonstrate the effects of the three different loss terms we used.
Fig.~\ref{fig:loss_graph} shows the change of loss values of generated pointcloud with different amount of missing data.
Both Chamfer loss $L_{CH}$ and GFV loss $L_{GFV}$ increase for a large amount of missing data. This is reasonable considering that we need to fill more information and make a larger change from the incomplete input as the ratio of missing data grows larger.
The $L_{D}$ is almost constant as the pre-trained generator synthesizes according to the learned distribution given the $z$ input.

\begin{figure}
  \centering
  \includegraphics[width=0.47\textwidth]{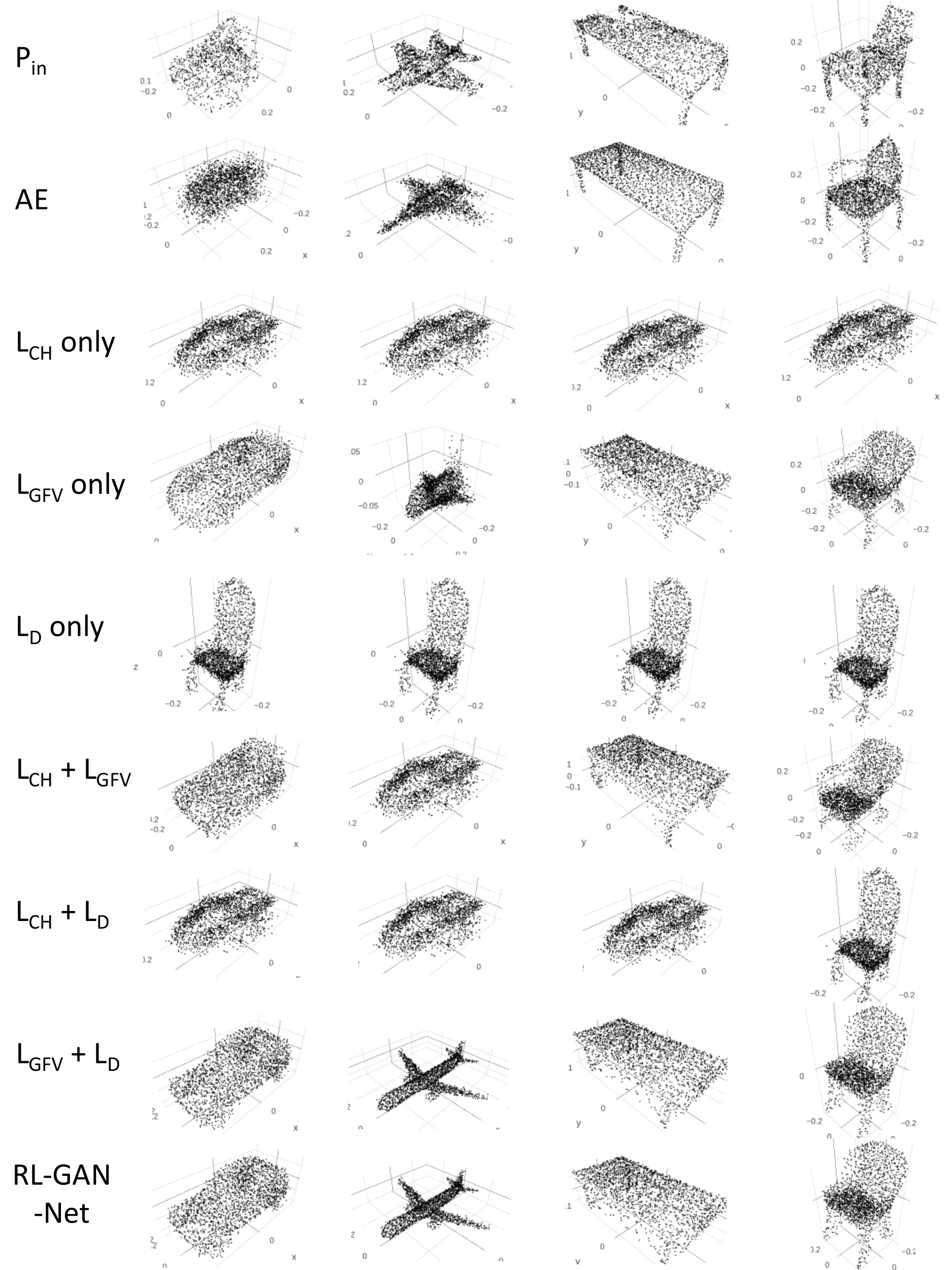}
  \caption{\small{{\bf Reward function analysis.} We tested the reward function with different combinations of losses. According to our analysis, the Chamfer loss cannot work alone to complete the shape but with the GFV loss our RL-GAN-Net can find the correct shape. The discriminator loss ensures that the completed shape is semantically meaningful.}}\label{fig:loss}
  \vspace{-1.0em}
\end{figure}

We also tested different combinations of loss functions for reward.
Fig.~\ref{fig:loss} shows the sample results with a shape per category.
While the Chamfer distance is the widely used metric to compare two shapes, the Chamfer loss was not very efficient when used alone.
This can be interpreted as the curse of dimensionality. 
While we need to semantically complete the 3D positions of 2048 points, the single number of Chamfer loss is not enough to inform the agent to find the correct control of the $l$-GAN.
On the other hand, the performance of GFV loss is impressive.
While details are often mismatched, the GFV loss alone enables the controller to find the correct semantic shape category from partial data.
This result agrees with the discussion in~\cite{panos}, where the latent space representation reduces the dimension and boost the performance of GAN.
However, the completed shape aligned better with the desired shape when combined with the Chamfer loss, which only shows its power when combined with GFV.
The discriminator loss is essential to create a realistic shape.
When discriminator loss is used alone, the RL agent creates a reasonable but unrelated shape, which is an expected behaviour considering the reward is simply encouraging a realistic shape. 
From the results, we conclude that all of the three loss terms are necessary for the RL agent to deduce the correct control of GAN.

\section{Conclusion and Future Work}

In this work, we presented a robust and real time point cloud shape completion framework using  the RL agent to control the generator.
Our primary motivation was to remove the costly and complex optimization process that is not real time and takes a minimum of 324 seconds to process a batch of inputs ~\cite{CMUshape}. Instead of optimizing the various combinations of loss functions, we have converted these loss functions into rewards. In doing so, our RL agent can complete the shape in approximately one millisecond. In addition, we present shape completion results with data with up to 70\% missing points. We show the superiority of our technique by demonstrating qualitative results. 

We have also presented a use case of our network for the classification problem. Being real time, RL-GAN-Net can be used to improve the performance of other point cloud processing networks. We demonstrated that our trained network raises the classification accuracy of PointNet from 50\% to 83\% for the data with 70\% missing points.  
With this work, we have demonstrated a hidden potential in RL-based techniques to effectively control the complex space of GAN.
An immediate extension is to apply the approach into closely related tasks such as image in-painting~\cite{inpainting}.
 \vspace{-0.95em}
\paragraph{Acknowledgements.} 
This work was supported by KIST institutional program [Project No. 2E29450] and the KAIST School of Electrical Engineering Graduate Fellowship. We are grateful to In So Kweon, Sunghoon Im, Arda Senocak from RCV lab, KAIST for insightful discussions.

\section{Implementation Details}

\begin{figure*}
    \centering
    \begin{subfigure}[b]{0.33\textwidth}
        \includegraphics[width=\textwidth]{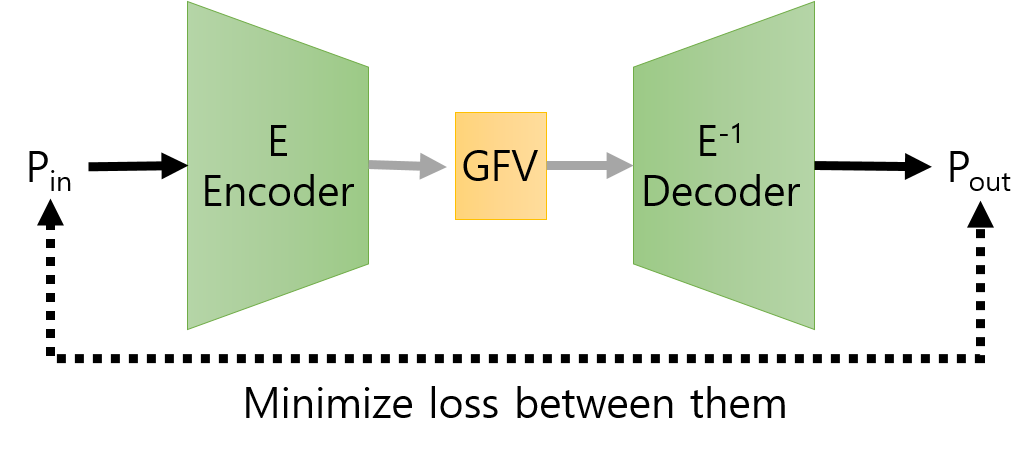}
        \caption{AE}
        \label{fig:AE}
    \end{subfigure}
    ~ 
    \begin{subfigure}[b]{0.32\textwidth}
        \includegraphics[width=\textwidth]{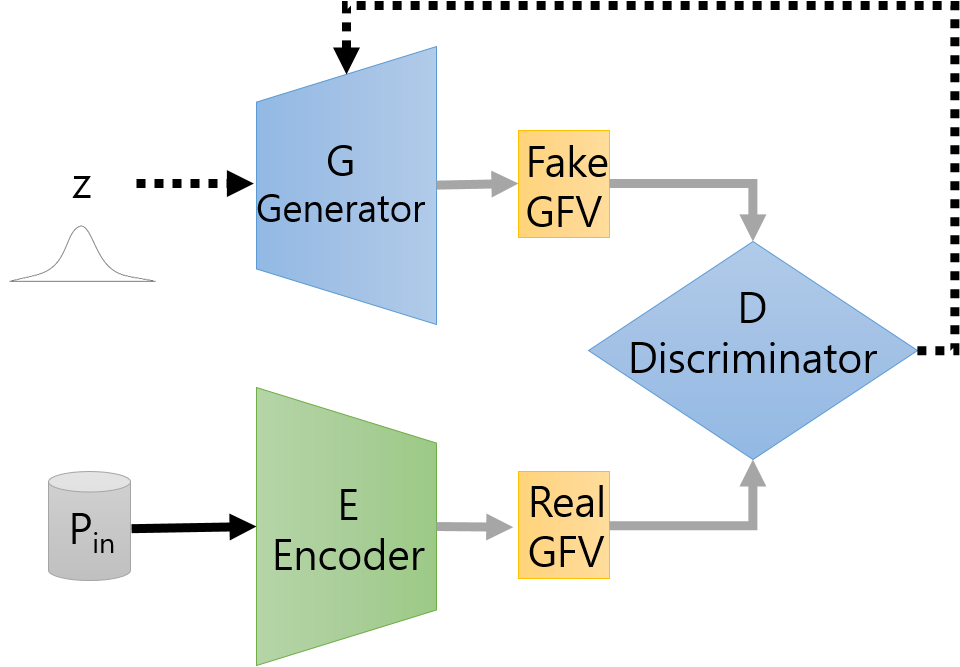}
        \caption{$l$-GAN}
        \label{fig:lGAN}
    \end{subfigure}
    ~ 
    \begin{subfigure}[b]{0.27\textwidth}
        \includegraphics[width=\textwidth]{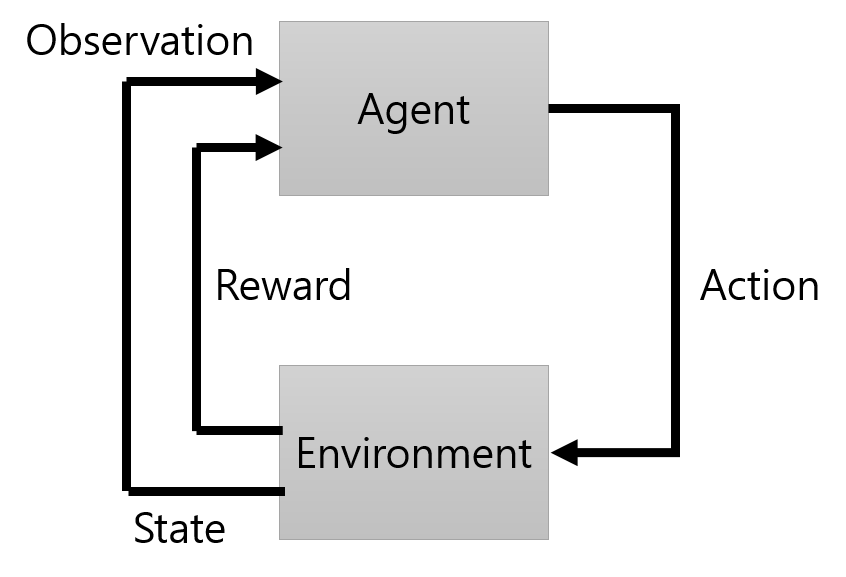}
        \caption{RL}
        \label{fig:RL}
    \end{subfigure}
    \caption{Network architecture of the three fundamental building blocks of RL-GAN-Net.}\label{fig:NN}
\end{figure*}

The suggested RL-GAN-Net architecture combines three fundamental building blocks as shown in Sec.~\textcolor{red}{3} of the main article. The general form of the architecture is provided in Fig.~\ref{fig:NN}.
In this section, we provide details of our implementation for each network.

\subsection{AE Details}
The AE is composed of an encoder that converts input points $P_{in}$ into GFV, and a decoder network that reverts GFV back to the point cloud domain, as shown in Fig.~\ref{fig:AE}. 
The input and output points are an unstructured list of 2048 3D coordinates that is sampled from the underlying 3D structure.  
The encoder network consists of five 1 D convolution layers with 64, 128, 128, 256, 128 channels respectively, while the decoder consists of FC layers 256, 256 and 6144 channels respectively. Each layer is followed by ReLu. The bottleneck size for AE is 128. We trained the AE to reduce the Chamfer distance (Eq.(\textcolor{red}{1}) of the main article) between the input and output point cloud. 
The Chamfer distance calculation is imported from the implementation\footnote[1]{https://github.com/lijx10/SO-Net} of Li et al~\cite{sonet}.

\subsection{$l$-GAN Details}
$l$-GAN is composed of the encoder block of AE and a generator and a discriminator, as shown in Fig.~\ref{fig:lGAN}. 
For the generator and the discriminator pair, we adapted the main architecture of the GAN from Zhang et al. \cite{selfgan} and applied in the latent space acquired by the AE.
The detailed network architecture for our modified $l$-GAN pipeline has been shown in Table~\ref{tb:gen} and~\ref{tb:disc} respectively. 
We trained the GAN using WGAN-GP~\cite{wgangp} adversarial loss with $\lambda_{gp}=10$. The total number of iterations was one million.
As a typical GAN training, we updated discriminator 5 times for every update of the generator. We used Adam optimizer~\cite{adam} with $\beta_1$=0.5 and $\beta_2$=0.9. The learning rate for both generator and discriminator was set to 0.0001. Batch size was set to 50 and number of workers were set to 2. We did not use any learning rate decay.  

We selected the dimension of $z$-vector to be 1. We did this to limit the dimensions of action space for the agent. All the experiments conducted were with a single dimension. We also tested with 6 and 32 dimensions but there was no change in the performance of the GAN or the agent in either case. Therefore we kept the dimension of the $z$-vector to 1.

We trained the GAN using the dataset generated by passing the ShapeNet point cloud dataset through the encoder of the AE. Therefore, the output dimension of our generator is the same as the bottleneck size of our AE (128). 

For $l$-GAN training, we adopted the self-attention GAN open source code\footnote[2]{https://github.com/heykeetae/Self-Attention-GAN} \cite{selfgan}.

\begin{table*}
\centering
\begin{tabular}{|c|ccc|cc|c|c|c|}
  \hline
  Name &  Kernel & Stride &  Padding &  InpRes &  OutRes & Input & Activation & Norm\\
  \hline
  convtr2d-layer1  & 4x4 & - & - & 50x32x1x1 & 50x256x4x4 & $z$-vector & ReLu & SN,BN\\
  \hline
  convtr2d-layer2  & 3x3 & 2 & 2 & 50x256x4x4 & 50x128x5x5 & convtr2d-layer1 & ReLu & SN,BN\\
  \hline
  convtr2d-layer3  & 3x3 & 2 & 2 & 50x128x5x5 & 50x64x7x7 & convtr2d-layer2 & ReLu & SN,BN\\
  \hline
  Self-Atten\cite{selfgan}  & -  & - & - & 50x1x7x7 & 50x1x7x7  & convtr2d-layer3 & -& -\\
 \hline
  convtr2d-last  & 2x2  & 2 & 1 & 50x64x7x7 & 50x1x12x12  & Self-Atten & -& -\\
  \hline
  reshape1  & 1x1  & - & - & 50x1x12x12 & 50x144  & convtr2d-last & -& -\\
  \hline
  convtrans1d  & 1x1  & - & - & 50x144 & 50x128  & convtr2d-last & -& -\\
  \hline
\end{tabular}
  \caption{ The network architecture of the generator. convtr2d = 2D transposed convolutional layer, convtrans1d = 1D transposed convolution, SN = spectral normalization\cite{selfgan} and BN = batch normalization  }\label{tb:gen}
\end{table*}

\begin{table*}
\centering
\begin{tabular}{|c|ccc|cc|c|c|c|}
  \hline
  Name &  Kernel & Stride &  Padding &  InpRes &  OutRes & Input & Activation & Norm\\
  \hline
  convtrans1d  & 1x1 & - & - & 50x128 & 50x144 & input& -& -\\
  \hline
  reshape1  & - & - & - & 50x144 & 50x12x12 & convtrans1d & -& -\\
  \hline
  conv2d-layer1  & 3x3 & 2 & 2 & 50x12x12 & 50x64x7x7 & reshape1 & ReLu & SN,BN\\
  \hline
  conv2d-layer2  & 3x3 & 2 & 2 & 50x64x7x7 & 50x128x5x5 & conv2d-layer1 & ReLu & SN,BN\\
  \hline
  conv2d-layer3  & 3x3 & 2 & 2 & 50x128x5x5 & 50x256x4x4 & conv2d-layer2 & ReLu & SN,BN\\
  \hline
  Self-Atten\cite{selfgan}  & - & - & - & 50x256x4x4 & 50x256x4x4 & conv2d-layer3 & - & -\\
  \hline
  conv2d-last  & 4x4  & - & - & 50x256x4x4 & 50x1  & Self-Attention & -& -\\
  \hline
\end{tabular}
  \caption{ The network architecture of the discriminator. conv2d = 2D convolutional layer, convtrans1d = 1D transposed convolution, SN = spectral normalization~\cite{selfgan} and BN = batch normalization  }\label{tb:disc}
\end{table*}

\subsection{RL Agent Details}
The third element of the basic architecture is RL. The basic RL framework is composed of an agent and the environment as in Fig.~\ref{fig:RL}. 
Among many possible variations of the RL agent, we used the actor-critic architecture to enable continuous control of the $l$-GAN.

\paragraph{Actor and Critic Architecture}
The actor and critic networks are chosen to be fully connected (FC) layers. The actor has four FC layers with 400, 400, 300, 300 neurons with ReLu activation for the first three layers and $\tanh$ for the last layer respectively. The input to the actor is a 128-dimensional GFV. The output is a single dimension $z$-vector. The critic also has four FC layers with 400, 432, 300, 300 neurons with ReLu activation for the first two layers respectively. 

\paragraph{Reward Function Hyper-parameter}
In Eq.(\textcolor{red}{5}) of the main article, the multiplicative weights of $w_{CH}$, $w_{GFV}$, and $w_D$ are  assigned to the corresponding loss functions. 
The weight values are chosen such that, when combined, the effects of individual terms are not out of proportion or dominant in any way. 
In other words, the total loss is within range for the RL agent to learn useful information for all of the terms of $L_{CH}$, $L_{GFV}$, and $L_D$.
For example, if the value for the Chamfer loss was approximately 1000 and the GFV loss was 10, then they are normalized by dividing by 100 and 1 respectively.  
After consulting the range of raw loss values of multiple trials, we set $w_{CH}=100$, $w_{GFV}=10.0$, and $w_D=0.01$ for all our experiments. 

The RL agent was adopted from the open source implementation of the DDPG algorithm.\footnote[3]{https://github.com/sfujim/TD3}.

\paragraph{Training Details}
The training of the agent can be divided into two parts. The first part is the collection of experience. The second part is the training of the actor and critic network in accordance with the DDPG algorithm as outlined in the previous work~\cite{ddpg}. 

For the first part, we refer the readers back to Fig. \textcolor{red}{3} of the main article. It shows the mechanism by which the replay buffer \textbf{R} is filled continuously with useful experiences. We fill the memory with one input at a time. This implies that the batch size for this case is one. Our task is episodic, which means that after each episode we collect a reward. The number of episodes is equal to the maximum number of allowed iterations. In each episode, the agent is allowed to take a single action after which the episode terminates. The sequences of state, action and reward tuples are then stored in the replay buffer. 

The second part, i.e., training the actor and critic in accordance with DDPG, is performed by keeping the batch size equal to one hundred. This means that a batch of a 100 memories from the replay buffer is picked randomly to train the actor and critic networks according to the DDPG algorithm. The evaluation of the policy was carried out after 5000 iterations. The number of dimensions of state is 128, which is basically the noisy GFV obtained by encoding the incomplete point cloud. The action dimension is determined by the dimension of the GAN's $z$-space, which is 1. The action space is kept to unity to achieve better performance by the agent. We also tested with 32 dimensions for $z$ space but it did not have any noticeable effect on the performance of GAN or the agent.

We list the parameter values used for the training with DDPG algorithm in Table~\ref{tb:param_RL}.

\begin{table}
\centering
\begin{tabular}{|c|c|}
  \hline
  Parameter & Value \\
  \hline
  maximum number of iterations & 1e6\\
  exploration noise & 0.1 \\
  batch size from \textbf{R} for the actor training & 100 \\
  discount $\gamma$ & 0.9 \\
  speed of target value updates $\tau$ & 0.005 \\
  noise added to policy during critic update & 0.2\\
  range to clip noise policy & 0.5 \\
  frequency for delayed policy update & 2 \\
  \hline
  \end{tabular}
  \caption{The parameter values used to train the RL agent.}
  \label{tb:param_RL}
  \end{table}

\section{Additional Results}
In this section, we provide enlarged images of the experiments in Sec.~\textcolor{red}{4} of the original document and include some additional results that were omitted due to the page limit.

\subsection{Shape Completion Results}
\begin{figure*}
  \centering
  \includegraphics[width=0.9\textwidth,trim={0 0.5cm 0 0},clip]{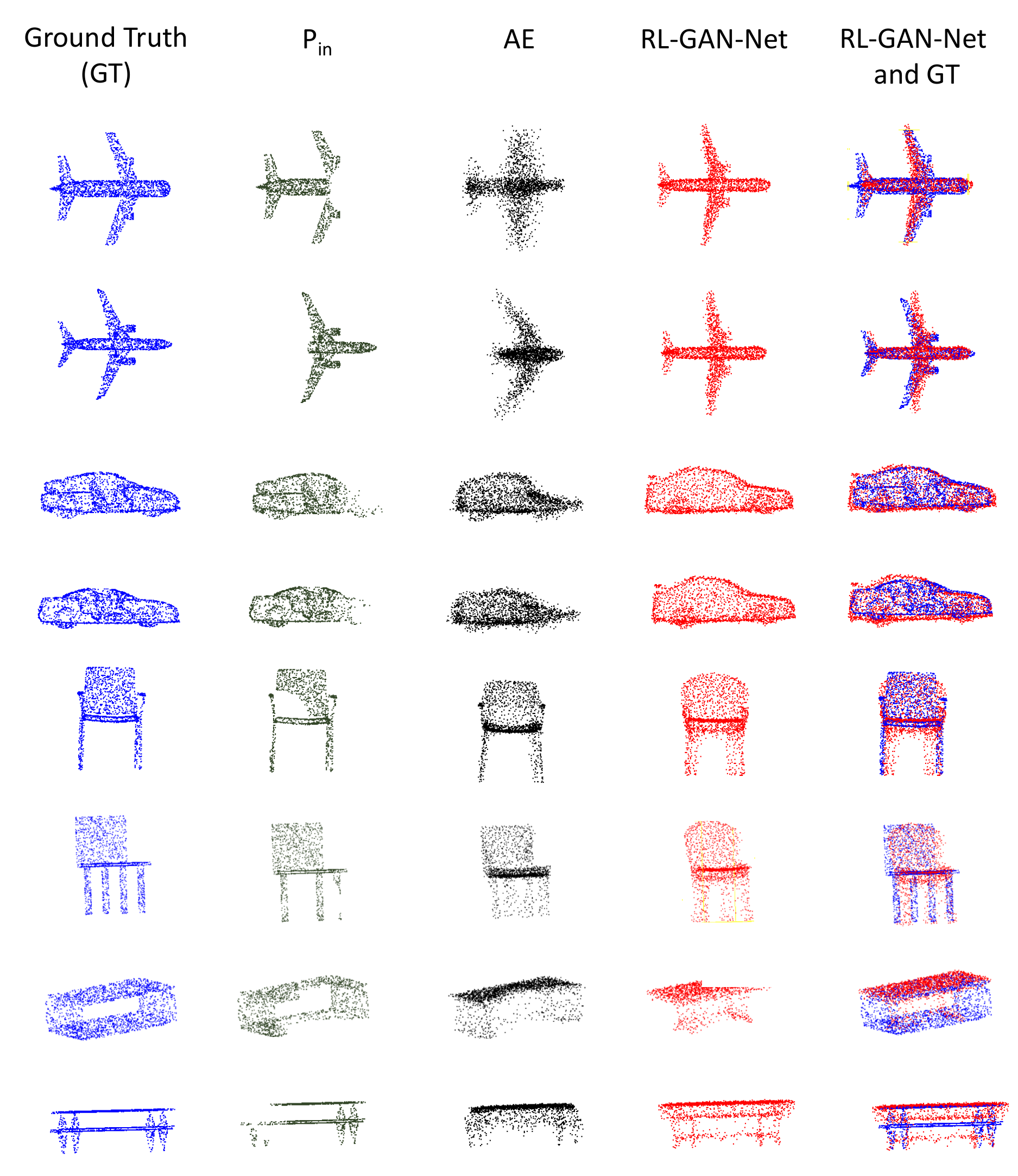}
  \caption{{ Qualitative results of point cloud shape completion missing 30\% of its original points.}}\label{fig:results30}
\end{figure*}

\begin{figure*}
  \centering
  \includegraphics[width=0.9\textwidth,trim={0 0.5cm 0 0},clip]{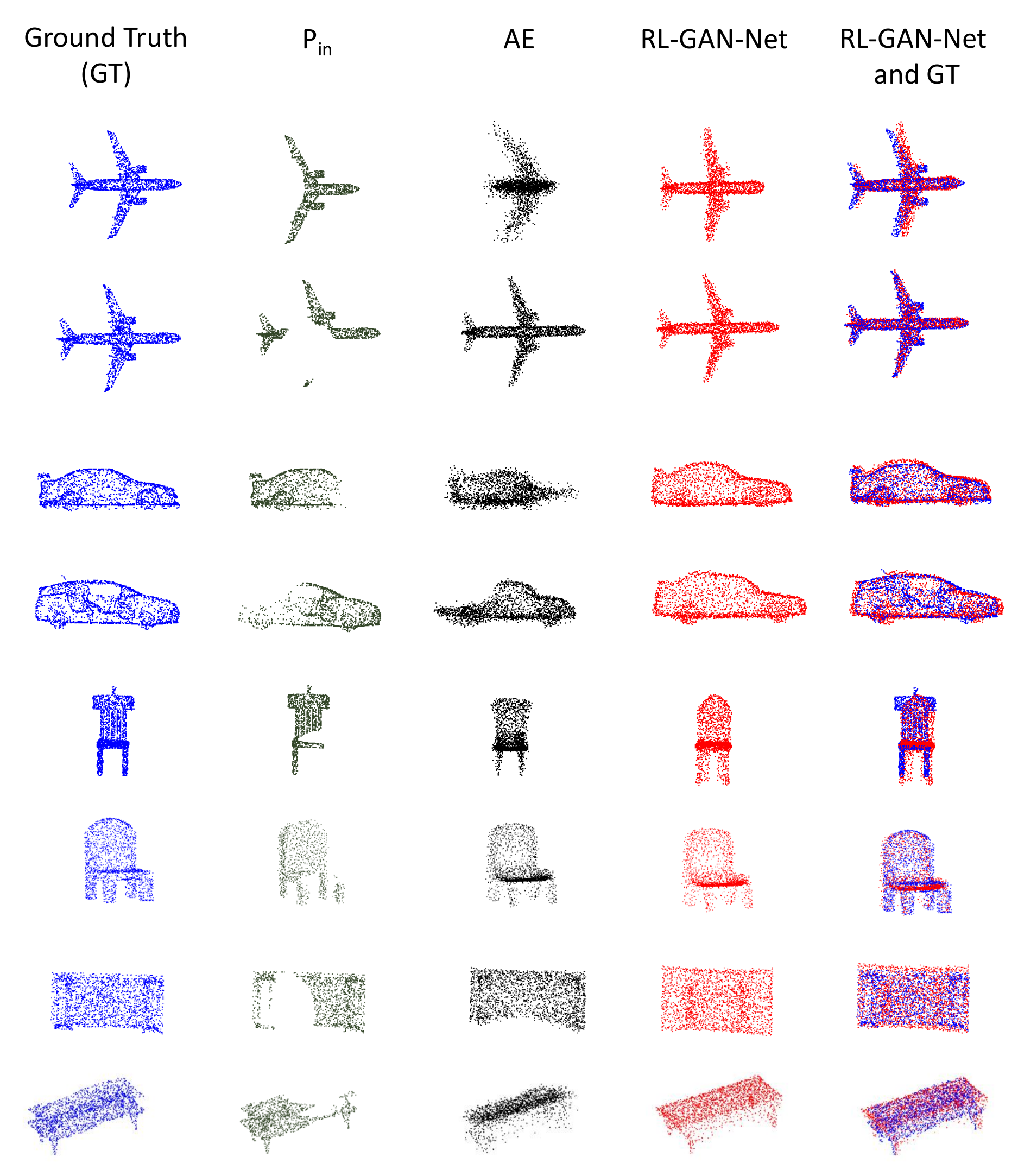}
  \caption{{ Qualitative results of point cloud shape completion missing 40\% of its original points.}}\label{fig:results40}
\end{figure*}

\begin{figure*}
  \centering
  \includegraphics[width=0.9\textwidth,trim={0 0.5cm 0 0},clip]{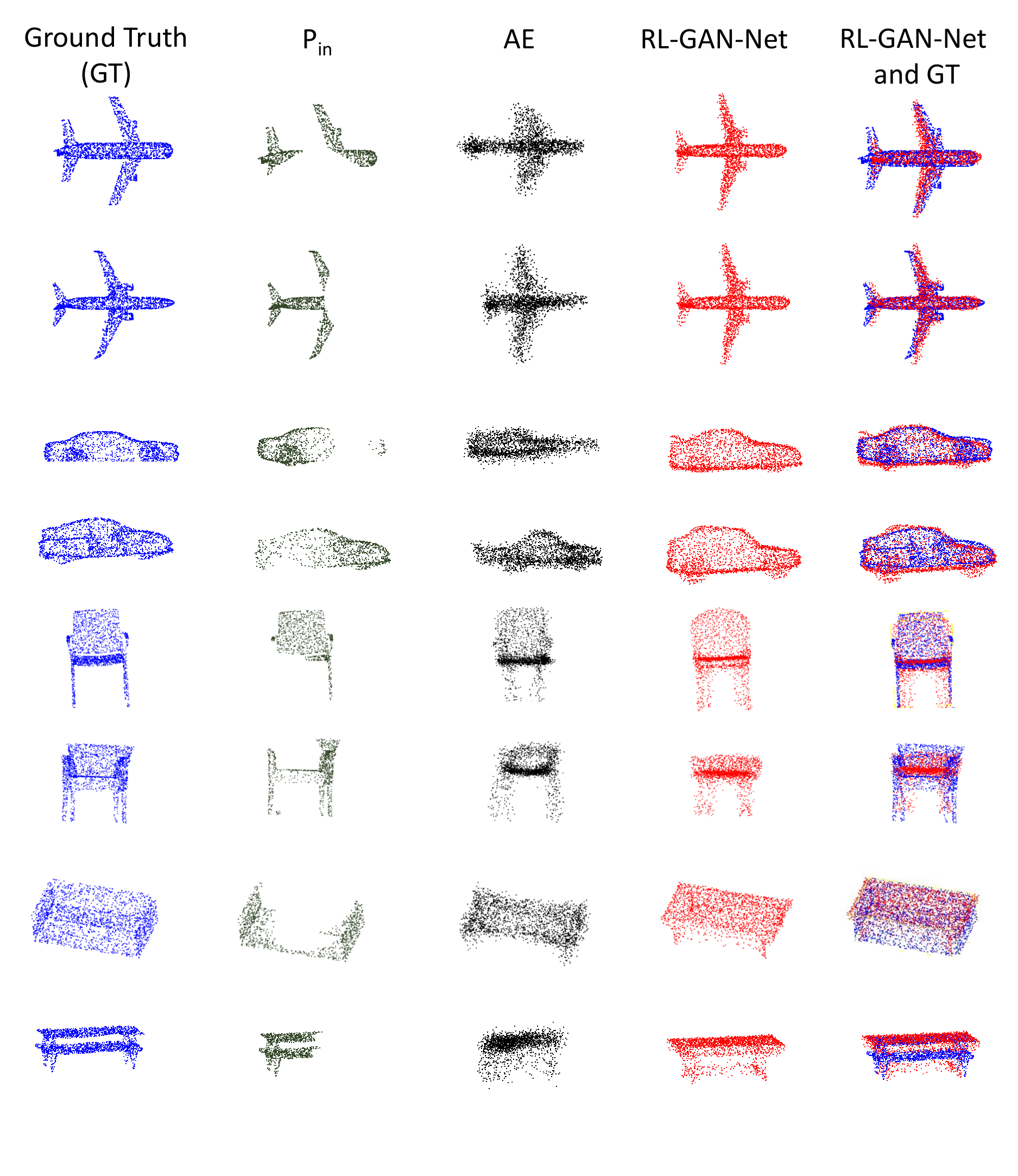}
  \caption{{ Qualitative results of point cloud shape completion missing 50\% of its original points.}}\label{fig:results50}
\end{figure*}

The examples of shape completion results for point cloud missing 20\% and 70\% of its original points are enlarged in Fig.~\ref{fig:results20} and Fig.~\ref{fig:results70}. In addition, we provide the examples of results for remaining data sets we used, which are missing 30\%, 40\% and 50\% of the original points as shown in Fig.~\ref{fig:results30},~\ref{fig:results40} and ~\ref{fig:results50} respectively. 
It is clear that the performance of our pipeline is prominent as the percentage of missing portion increases.

\begin{figure*}
  \centering
  \includegraphics[width=0.9\textwidth,trim={0 0.5cm 0 0},clip]{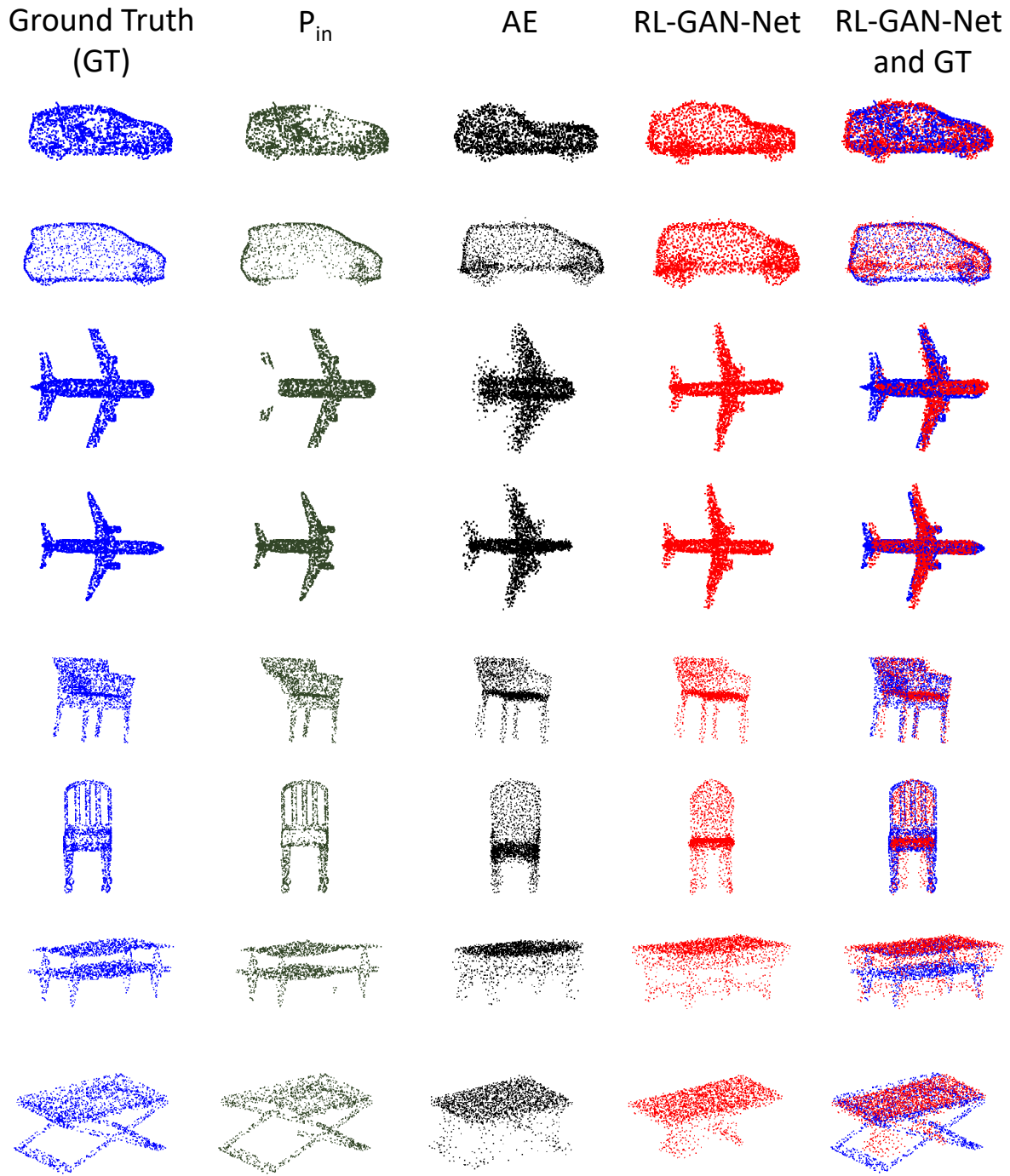}
  \caption{{ Qualitative results of point cloud shape completion missing 20\% of its original points.} }\label{fig:results20}
\end{figure*}

\begin{figure*}
  \centering
  \includegraphics[width=0.9\textwidth,trim={0 0.5cm 0 0},clip]{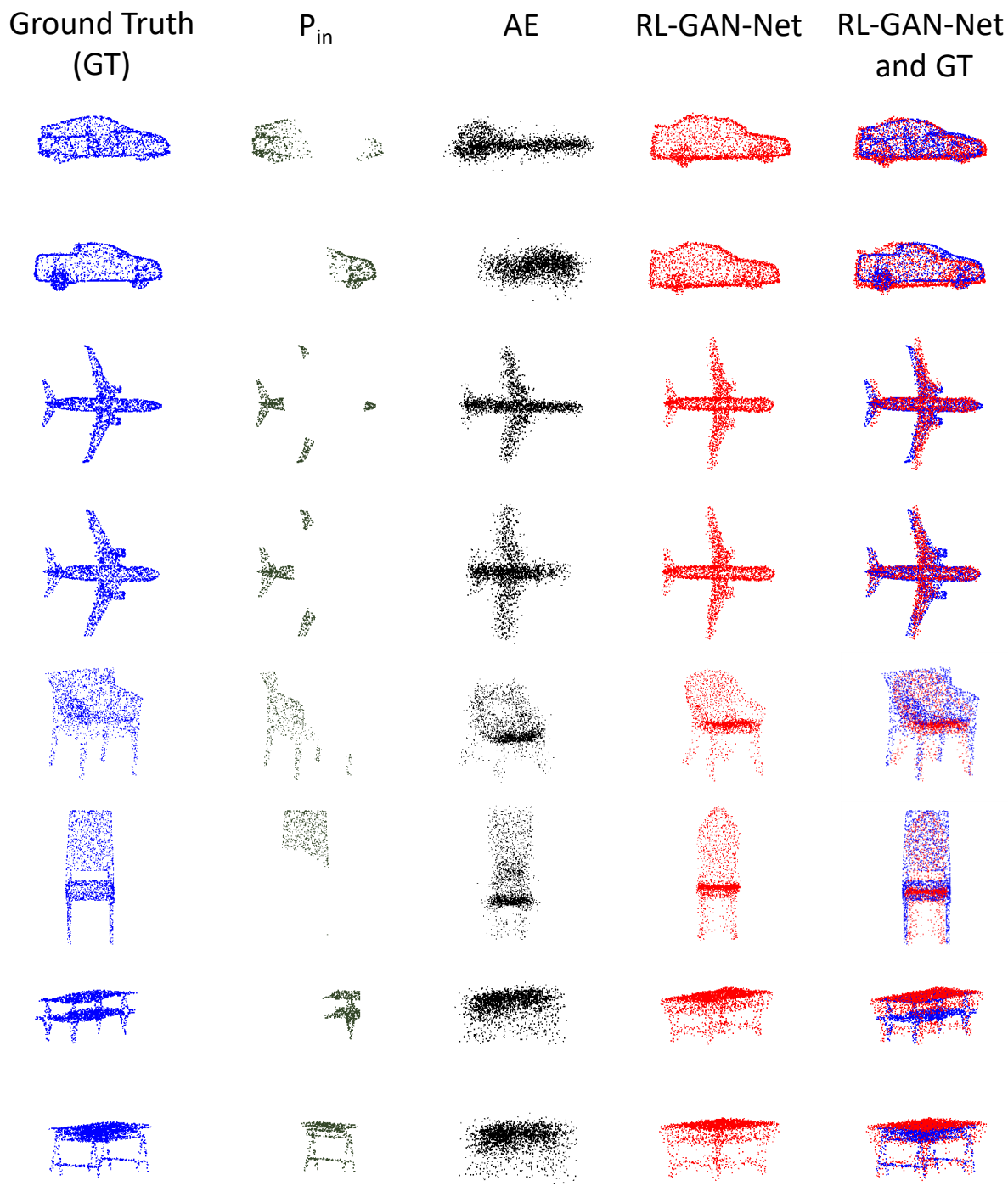}
  \caption{{ Qualitative results of point cloud shape completion given input data missing 70\% of  its original points.}}\label{fig:results70}
\end{figure*}

\subsection{Robustness Results}
\begin{figure*}
  \centering
  \includegraphics[width=0.9\textwidth,trim={0 0.5cm 0 0},clip]{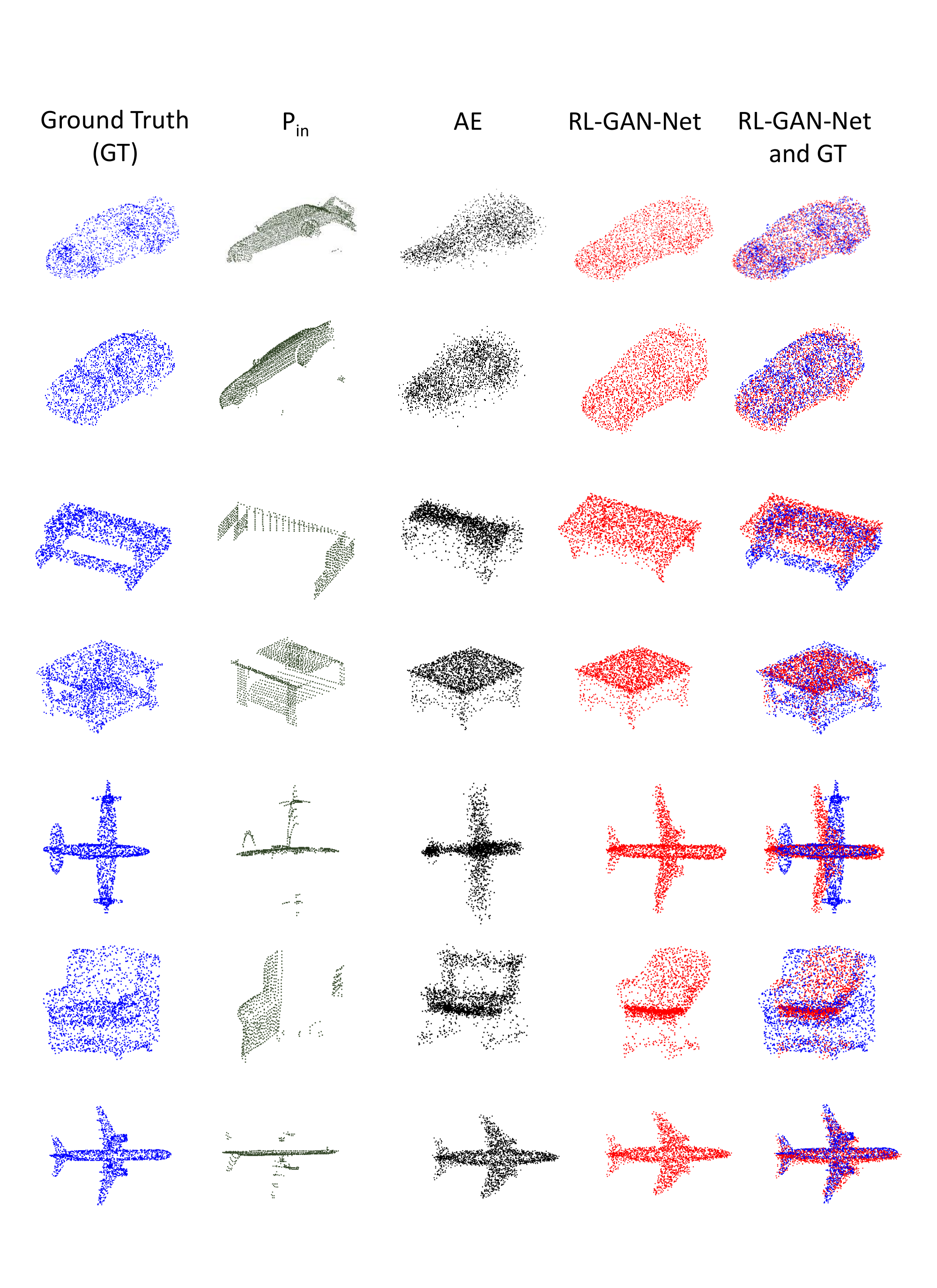}
  \caption{\textbf{Robustness test.} We applied our algorithm to the point cloud data provided by~\cite{daiEPN}. This figure shows examples of shape completion results with the raw scan data provided.}\label{fig:robustnojitter}
\end{figure*}

\begin{figure*}
  \centering
  \includegraphics[width=0.9\textwidth,trim={0 0.5cm 0 0},clip]{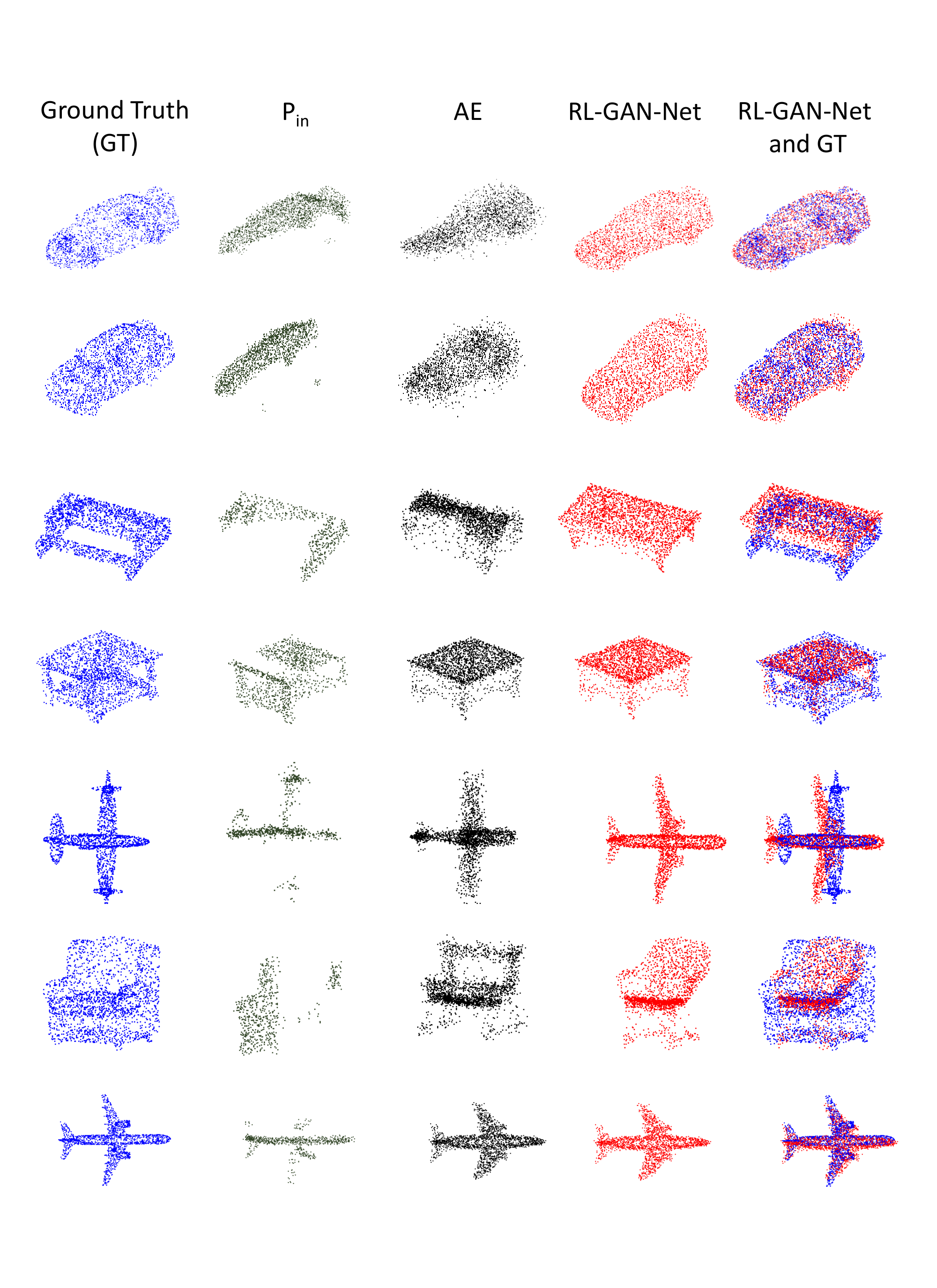}
  \caption{\textbf{Robustness test.} We applied our algorithm to the point cloud data provided by~\cite{daiEPN}. This figure shows results when we added zero-mean Gaussian noise with standard deviation 0.01 (clipped at 0.05). }\label{fig:robustjitter}
\end{figure*}

The robustness test results with the different dataset provided by Dai et al~\cite{daiEPN} are included in  Fig.~\ref{fig:robustnojitter} and Fig.~\ref{fig:robustjitter}.
Our result is almost not affected by the jitter, and the completed shape is semantically similar to its original shape.

For the cases where there is no jitter, we also include the completion results of Dai et al~\cite{daiEPN}.
Their approach works in a different domain (voxel grid) but we are including a comparison as there is not much prior work in point cloud space.
To briefly describe, their approach used an encoder-decoder network in $32^3$ voxel space followed by an analytic patch-based completion in $128^3$ resolution.
Their results of both resolutions are available as distance function format.
We converted the distance function into a surface representation using the MATLAB function {\it isosurface} as they described, and uniformly sampled 2048 points to compare with our results.
By comparing against the ground truth model, ours is superior to their approach in terms of the Chamfer distance as shown in Table \ref{tb:dai}. It should be noted here that the Chamfer distance between the input and ground truth is comparable to autoencoder. This is expected because the dataset provided by Dai et al.\cite{daiEPN} does not have any drastic level of incompletion in many cases. We also refer the reader back to the Fig.\textcolor{red}{5a} of the main article where clearly the Chamfer distance of the input point cloud compared to the ground truth was even lower than an AE for missing data percentages less than forty.

We present the qualitative visual comparison in Fig.~\ref{fig:angela}.
The results of encoder-decoder based network (refered as Voxel $32^3$ in the figure) are smoother than point clouds processed by AE as the volume accumulation compensates for random noise. However, the approach is limited in resolution and washes out the local details. 
Even after the patch-based synthesis in $128^3$ resolution, the details they could recover are limited.
On the other hand, our approach robustly preserves semantic symmetries and completes local details in challenging scenarios.
It should be noted that we used only scanned point data but did not incorporate the additional mask information, which they utilized.

\begin{table}
\centering
\begin{tabular}{|c|c|c|c|c |}
  \hline
 Input & V $32^3$ & V $128^3$ & AE & RL-GAN-Net \\
  \hline
  0.0688 & 0.169& 0.162 & 0.0531 & 0.0690 \\
  \hline
  \end{tabular}
  \caption{The Chamfer distance compared to the ground truth. There are two volumetric approaches compared against two point cloud based approach. V $32^3$ is the results using encoder-decoder based network in voxel space at the resolution of 32 per dimension, and V $128^3$ shows the distance after the full pipeline including patch-based synthesis in~\cite{daiEPN}. AE and RL-GAN-Net are the point cloud based approaches of~\cite{panos} and ours.}
  \label{tb:dai}
  \end{table}

\begin{figure*}
  \centering
  \includegraphics[width=1.0\textwidth,trim={0 0.5cm 0 0},clip]{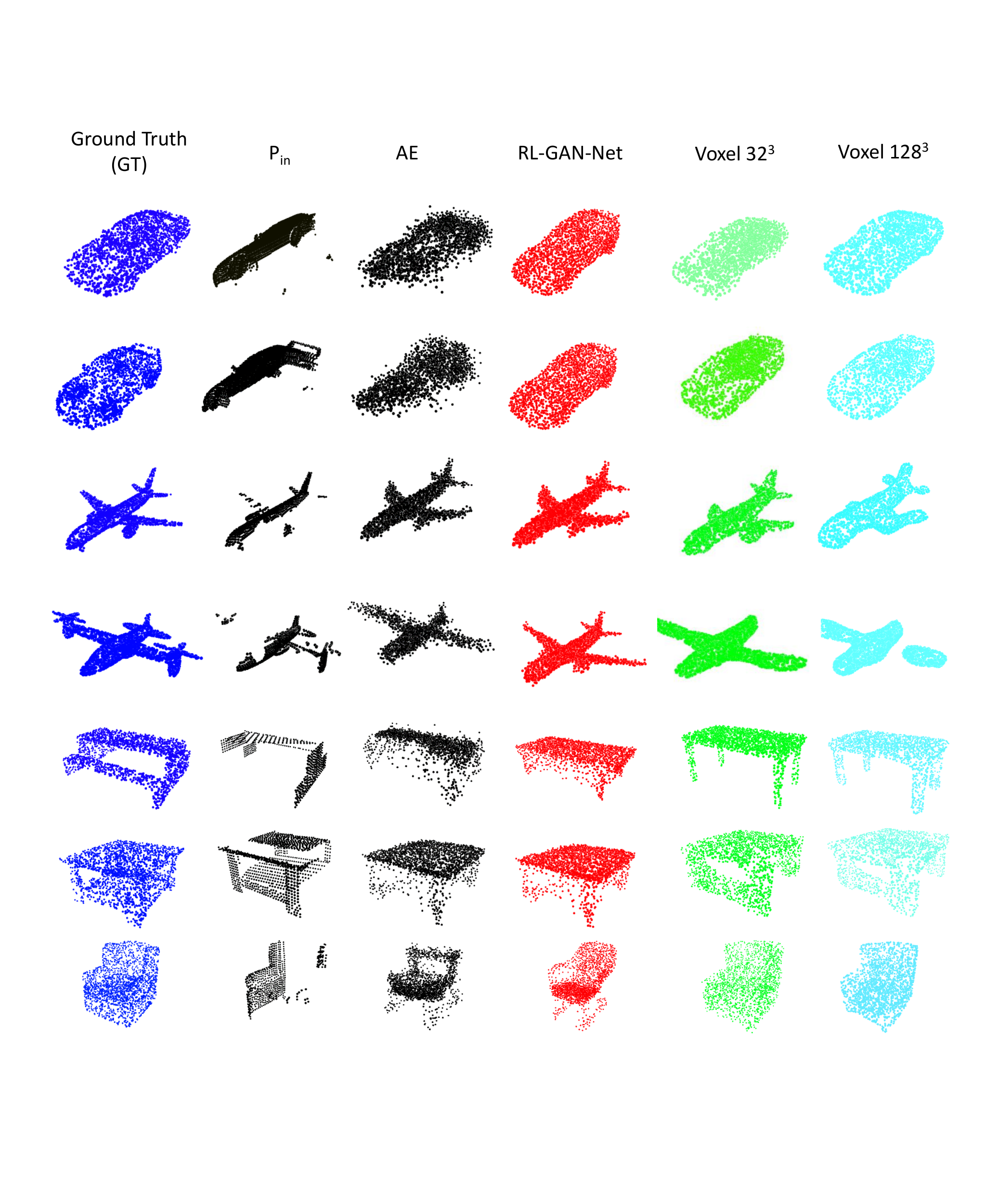}
  \caption{\textbf{Performance Comparison.} Comparison of RLGAN-Net vs Dai et al.\cite{daiEPN} for their $32^3$ and $128^3$ resolution results. We converted their distance function output to point cloud domain. It should be noted that they additionally have mask  information whereas we operate directly on the scanned points only.}\label{fig:angela}
\end{figure*}

\subsection{Classifier Details}
\begin{figure*}
    \centering
    \begin{subfigure}[b]{0.7\textwidth}
        \includegraphics[width=\textwidth]{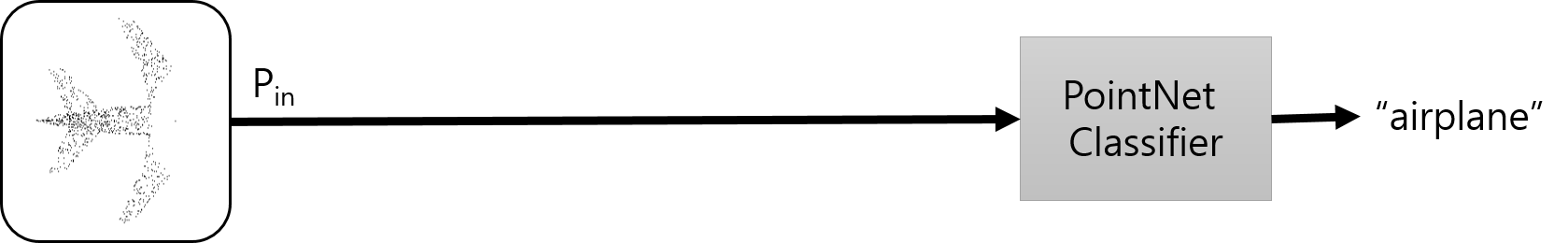}
        \caption{Vanilla PointNet Classifier}
        \label{fig:classraw}
    \end{subfigure}\vspace{1cm}
    ~ 
    \begin{subfigure}[b]{0.7\textwidth}
        \includegraphics[width=\textwidth]{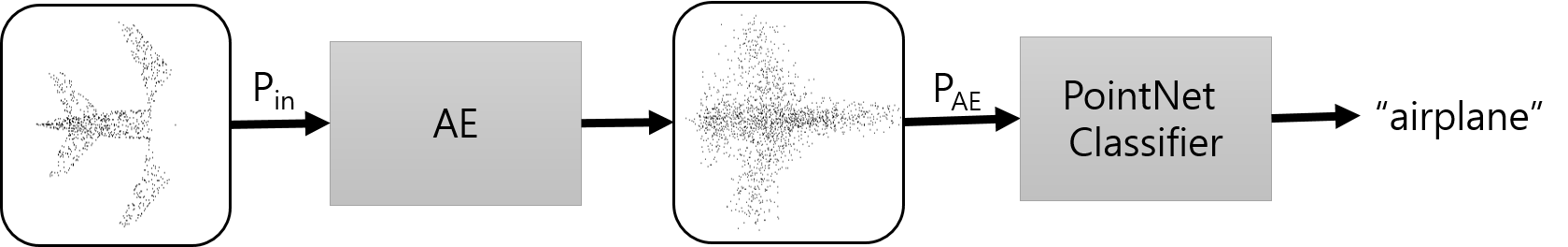}
        \caption{AE + PointNet Classifier}
        \label{fig:classAE}
    \end{subfigure}\vspace{1cm}
    ~ 
    \begin{subfigure}[b]{0.7\textwidth}
        \includegraphics[width=\textwidth]{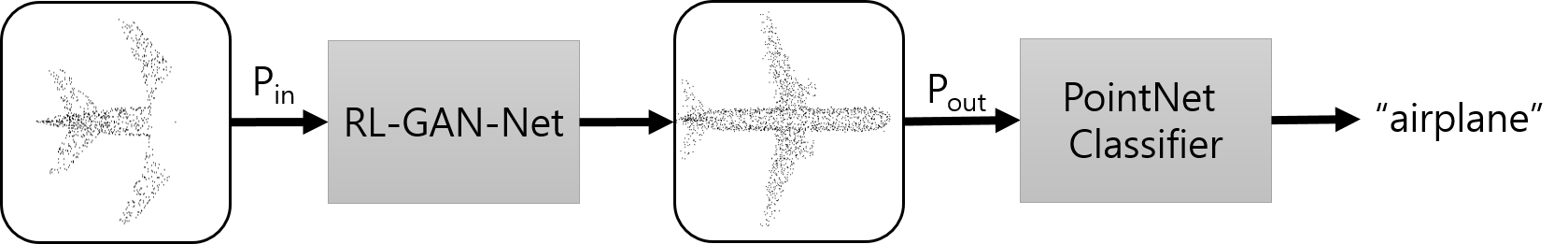}
        \caption{RL-GAN-Net + PointNet Classifier}
        \label{fig:classRL}
    \end{subfigure}\vspace{1cm}
    \caption{The variations of network architecture for point cloud classification with missing data}\label{fig:class}
\end{figure*}

We have trained the PointNet \cite{pointnet} classifier to distinguish the four categories that our RL-GAN-Net was trained on. 
After training, it classifies the shapes with 99.36\% of accuracy on the full data set with a complete point cloud. 
At test time, we consider the three scenarios as shown in Fig.~\ref{fig:class}, namely using the raw partial input, and using the shapes processed and completed by AE and RL-GAN-Net. 
The three pipelines are tested with the incomplete point cloud dataset for classification accuracy. 
For the cases that more than 30\% of the original shape data is missing, the classification accuracy is boosted when the shapes are pre-processed with shape completion pipeline. And our suggested pipeline is superior to AE and robust to large missing regions.
The detailed classification results are shown in Table~\ref{tb:classify}.

\begin{table*}
\centering
\begin{tabular}{|c|c|c|c|c|c|}
  \hline
  Network &  20\% & 30\% &  40\% &  50\% &  70\% \\
  \hline
  PointNet\cite{pointnet}(Fig.~\ref{fig:classraw})  & {\bf 98.6} & 95.4 & 85.2 & 73.9 & 50.2 \\
  \hline
  AE\cite{panos} + PointNet (Fig.~\ref{fig:classAE})  & 98.5 & 96.0 & 89.6 & 80.4 & 69.6 \\
  \hline
  RL-GAN-Net (vanilla) + PointNet (Fig.~\ref{fig:classRL})  & 97.7 & 96.7 & 95.0 & 92.7 & 82.5\\
  \hline
  RL-GAN-Net (hybrid) + PointNet (Fig.~\ref{fig:classRL})  & 98.1 & {\bf 97.2} & {\bf 95.5} & {\bf 93.3} & {\bf 83.8}\\
  \hline
  
\end{tabular}
  \caption{Classification accuracy of point cloud input processed by RL-GAN-Net compared to vanilla and AE for various percentage of missing data points }\label{tb:classify}
\end{table*}


{\small
\bibliographystyle{ieee_fullname}
\bibliography{egbib}
}

\end{document}